\documentclass{article}

\usepackage[final,nonatbib]{neurips_2025}

\usepackage[utf8]{inputenc} %
\usepackage[T1]{fontenc}    %
\usepackage{hyperref}       %
\usepackage{url}            %
\usepackage{booktabs}       %
\usepackage{amsfonts}       %
\usepackage{nicefrac}       %
\usepackage{microtype}      %
\usepackage{xcolor}         %
\usepackage[table]{xcolor}
\definecolor{mygray}{RGB}{240,240,240}      %
\definecolor{myblue}{RGB}{230,240,255}
\definecolor{mygreen}{RGB}{240,255,240}
\usepackage{multirow}
\usepackage{floatrow}
\usepackage{wrapfig}
\usepackage{graphicx}
\usepackage{makecell}
\usepackage{enumitem}
\usepackage{amsmath}
\usepackage{amsthm}
\usepackage{subcaption}
\usepackage{caption}
\usepackage{algorithm}
\usepackage{algorithmic}

\newtheorem{remark}{Remark}

\title{Delving into Cascaded Instability:\\ A Lipschitz Continuity View on Image Restoration and Object Detection Synergy}


\author{%
  Qing Zhao\textsuperscript{1} \quad
  Weijian Deng\textsuperscript{2} \quad
  Pengxu Wei\textsuperscript{1,5}\thanks{Corresponding Author. \quad Code available: \url{https://github.com/diasuki/LR-YOLO}} \\
  \textbf{Ziyi Dong\textsuperscript{1}} \quad
  \textbf{Hannan Lu\textsuperscript{3}} \quad
  \textbf{Xiangyang Ji\textsuperscript{4}} \quad
  \textbf{Liang Lin\textsuperscript{1,5}}\\[3pt]
  \textsuperscript{1}Sun Yat-sen University \quad
  \textsuperscript{2}Australian National University \\
  \textsuperscript{3}Harbin Institute of Technology \quad
  \textsuperscript{4}Tsinghua University \quad 
  \textsuperscript{5}Peng Cheng Laboratory \quad 
  \\[3pt]
  \{zhaoq78, dongzy6\}@mail2.sysu.edu.cn, 
  dengwj16@gmail.com,\\
  weipx3@mail.sysu.edu.cn,
  luhannan@hit.edu.cn,
  xyji@tsinghua.edu.cn,
  linliang@ieee.org
}

\begin{document}

\maketitle

\begin{abstract}
To improve detection robustness in adverse conditions (e.g., haze and low light), image restoration is commonly applied as a pre-processing step to enhance image quality for the detector. However, the functional mismatch between restoration and detection networks can introduce instability and hinder effective integration---an issue that remains underexplored.
We revisit this limitation through the lens of Lipschitz continuity, analyzing the functional differences between restoration and detection networks in both the input space and the parameter space. Our analysis shows that restoration networks perform smooth, continuous transformations, while object detectors operate with discontinuous decision boundaries, making them highly sensitive to minor perturbations. This mismatch introduces instability in traditional cascade frameworks, where even imperceptible noise from restoration is amplified during detection, disrupting gradient flow and hindering optimization.
To address this, we propose Lipschitz-regularized object detection (LROD), a simple yet effective framework that integrates image restoration directly into the detector’s feature learning, harmonizing the Lipschitz continuity of both tasks during training. We implement this framework as Lipschitz-regularized YOLO (LR-YOLO), extending seamlessly to existing YOLO detectors. Extensive experiments on haze and low-light benchmarks demonstrate that LR-YOLO consistently improves detection stability, optimization smoothness, and overall accuracy.
\end{abstract}

\section{Introduction}

Adverse imaging conditions introduce challenges for object detection by causing various image degradations, including reduced contrast, blurred edges, and obscured object boundaries.
A typical way to alleviate this issue is to employ image restoration as a pre-processing step, aiming to improve image quality before detection~\cite{liu2022image,sun2022rethinking,kalwar2023gdip}. However, its effectiveness is limited by the functional mismatch between restoration and detection networks. 
This inconsistency can introduce instability, where imperceptible noise introduced during restoration is amplified during detection, leading to unreliable predictions~\cite{li2023detection, wu2024unsupervised}. Moreover, the underlying differences between these tasks remain underexplored, hindering opportunities for better integration and enhanced robustness.
To bridge this gap, understanding their functional behaviors is crucial for achieving effective synergy. To this end, we analyze the conventional Image Restoration\(\rightarrow\)Object Detection cascade framework through the lens of Lipschitz continuity, focusing on two aspects: the input space and the parameter space.

\begin{figure}[t!]
    \centering
    \begin{subfigure}[t]{0.24\textwidth}
        \centering
        \includegraphics[width=\linewidth]{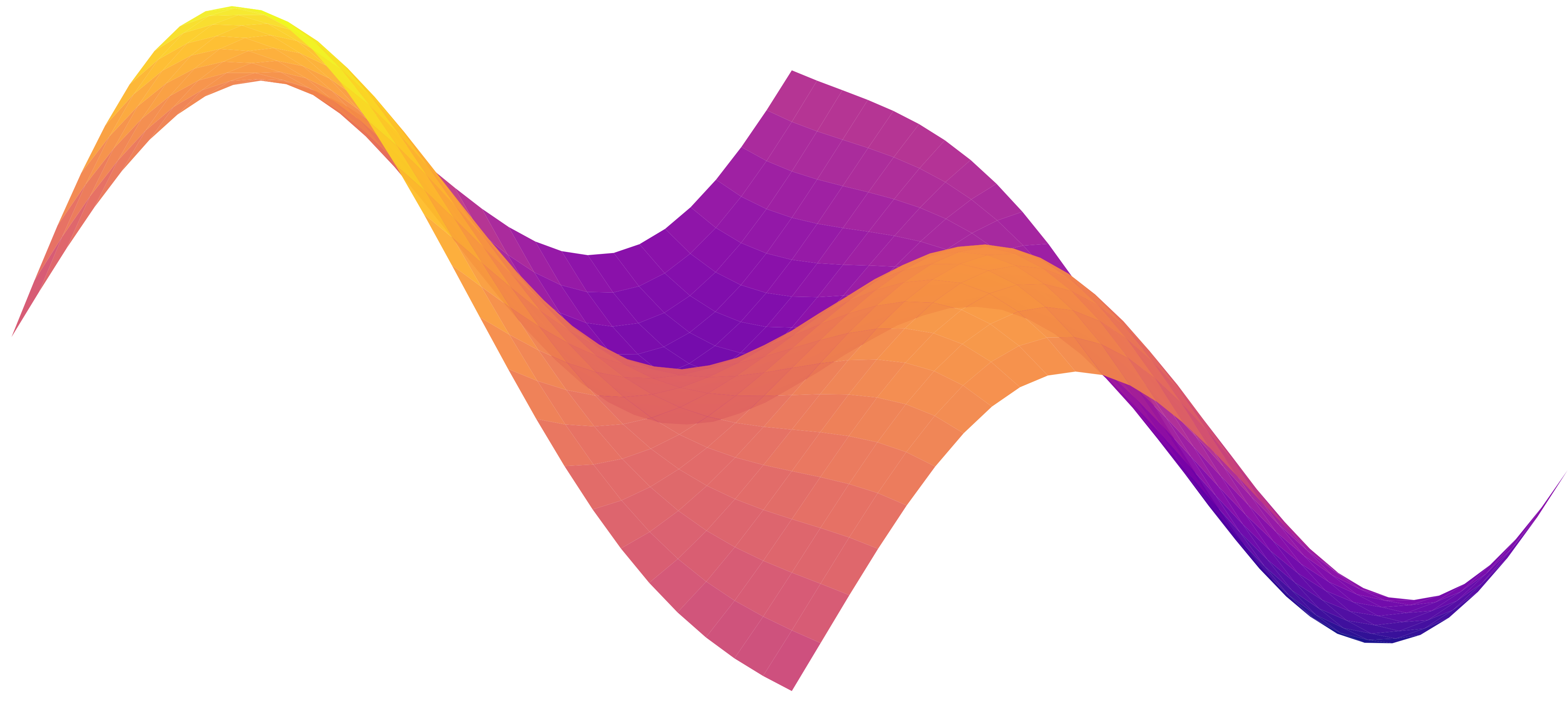}
        \caption{Image Restoration}
    \end{subfigure}
    \hfill
    \begin{subfigure}[t]{0.24\textwidth}
        \centering
        \includegraphics[width=\linewidth]{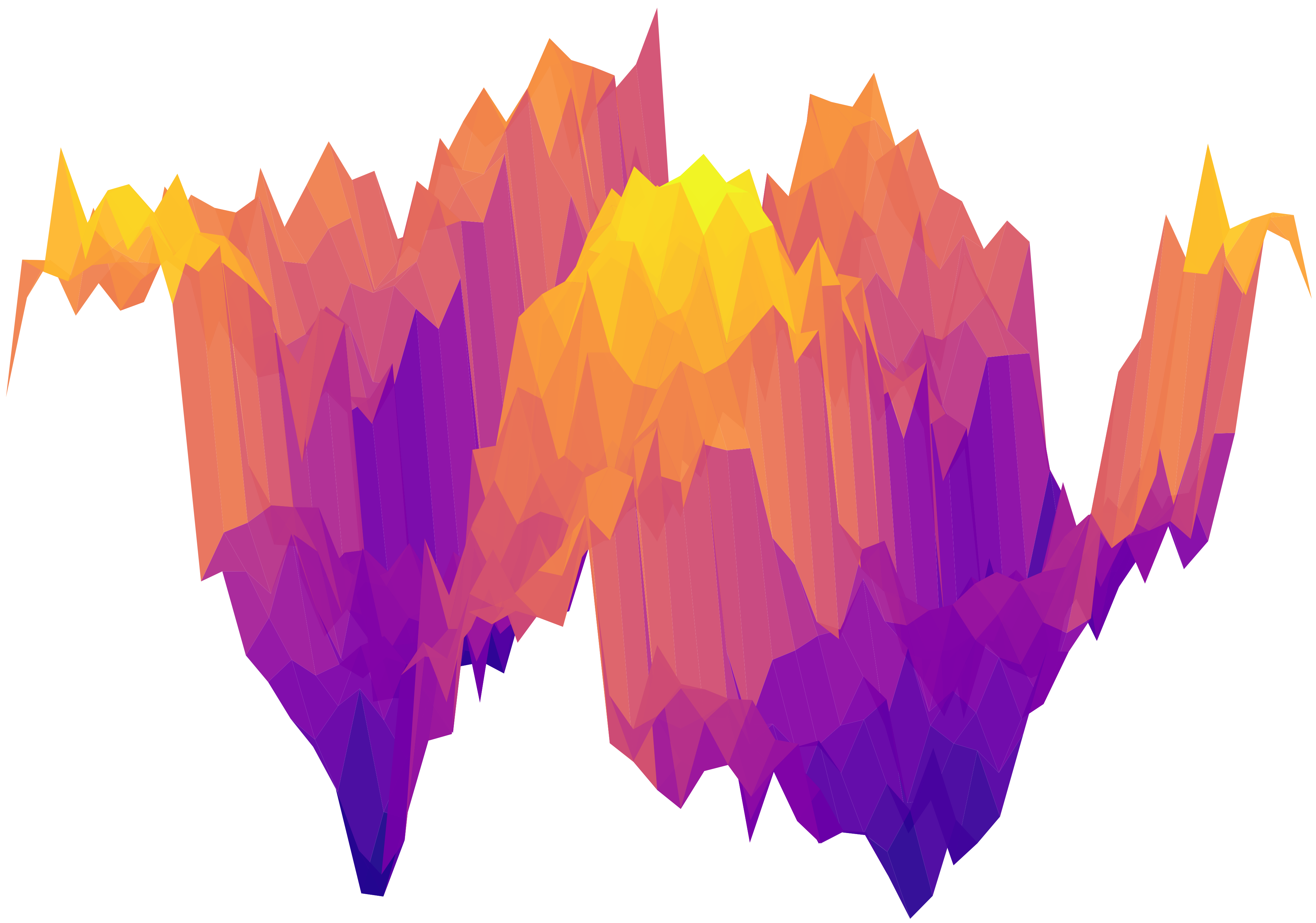}
        \caption{Object Detection}
    \end{subfigure}
    \hfill
    \begin{subfigure}[t]{0.24\textwidth}
        \centering
        \includegraphics[width=\linewidth]{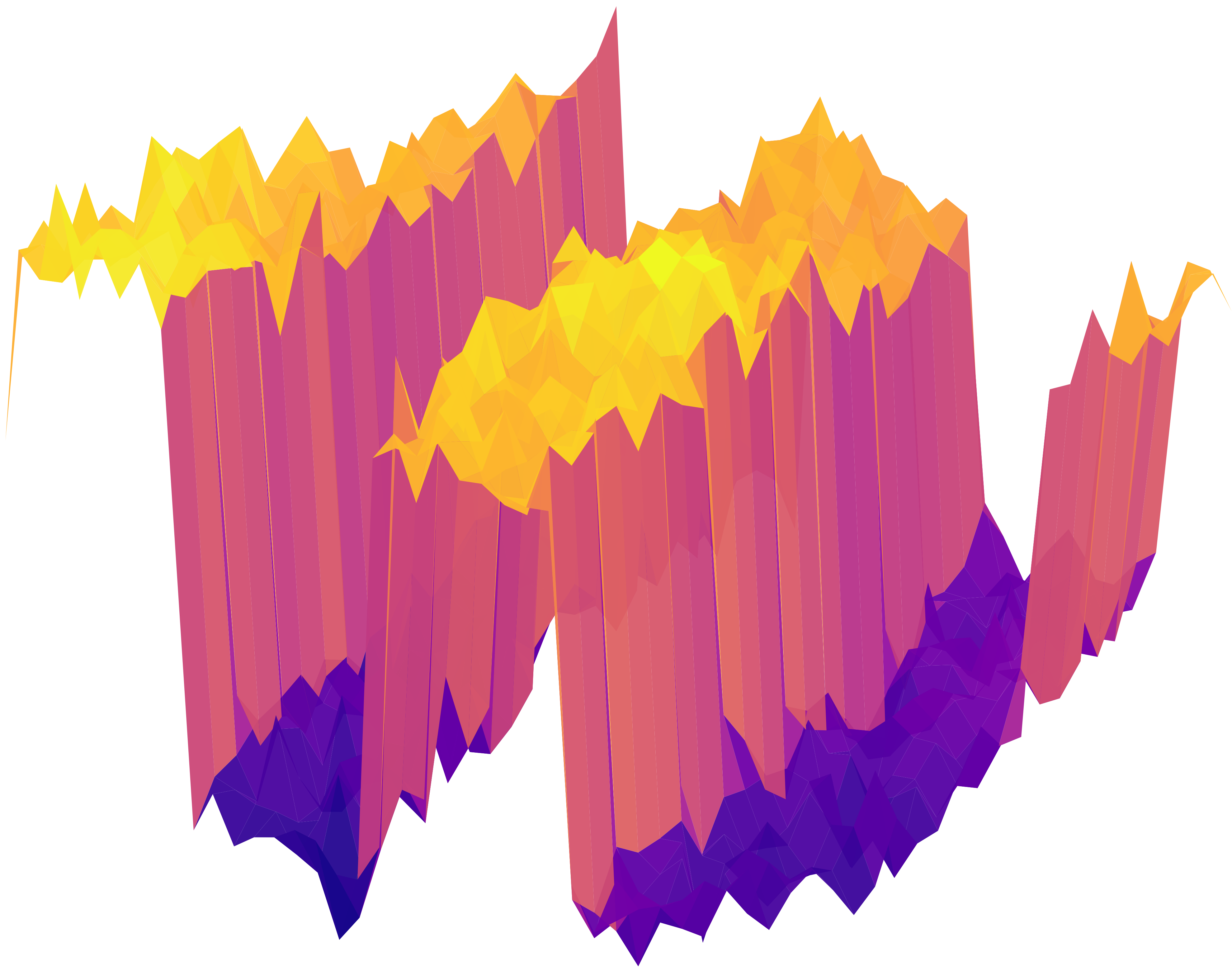}
        \caption{Cascade}
    \end{subfigure}\hfill
    \begin{subfigure}[t]{0.24\textwidth}
        \centering
        \includegraphics[width=\linewidth]{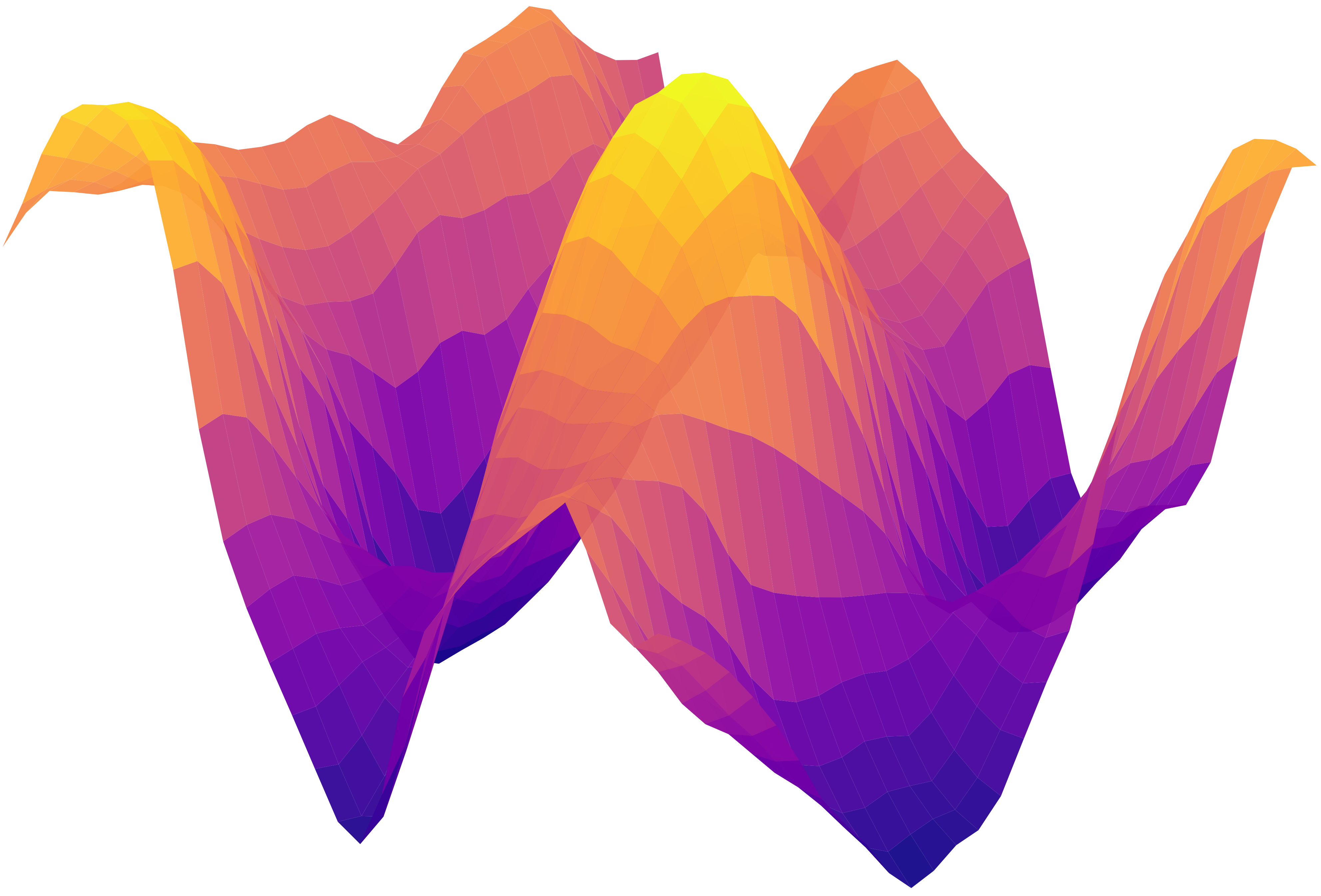}
        \caption{Ours}
    \end{subfigure}

    \caption{
    \textbf{Visualization of network functional behaviors under input perturbations.}
(a) Image Restoration networks exhibit smooth, continuous mappings, where input changes lead to gradual adjustments.
(b) Object Detection networks display sharp discontinuities due to abrupt decision boundaries in classification and bounding box regression.
(c) Cascade frameworks (Image Restoration \(\rightarrow\) Object Detection) magnify instability, resulting in fragmented and non-smooth behavior.
(d) Our method integrates low-Lipschitz image restoration into the feature learning of high-Lipschitz object detection, promoting smoother transitions and enhanced stability.
    }
    \label{fig:motivation}
\end{figure}

From the input space perspective, we leverage the concept of Lipschitz continuity, which characterizes the sensitivity of a model’s output to input perturbations~\cite{bubeck2021universal}. Networks with lower Lipschitz constants exhibit smoother, more predictable changes, while higher constants indicate heightened sensitivity and instability. By computing the Jacobian norm~\cite{khromov2023some} with respect to haze density variations, we observe that the Lipschitz constant of the object detection network is nearly an order of magnitude larger than that of the restoration network, highlighting its substantially lower smoothness.
This disparity highlights the differences in their functional behaviors. Restoration networks exhibit smooth, continuous mappings, where small input perturbations result in gradual and predictable adjustments to the restored image. This smoothness stems from pixel-wise processing that consistently enhances local regions and propagates changes smoothly across the image. In contrast, object detection networks are inherently discontinuous, characterized by sharp decision boundaries in classification and bounding box regression. Even minor pixel-level changes can cause abrupt shifts in class predictions or bounding box coordinates, reflecting non-smooth, step-like transitions in the output. This sharp contrast in behavior contributes to instability when the two networks are cascaded.
To further illustrate this disparity, we visualize the functional behaviors in Figure~\ref{fig:motivation} (a) and (b), where the smooth transitions of restoration sharply contrast with the abrupt shifts observed in detection. This inconsistency introduces instability when the two networks are cascaded: imperceptible noise introduced during restoration can be amplified during detection, resulting in overall non-smooth behavior in the cascade framework, as shown in Figure~\ref{fig:motivation}~(c).

To further understand the instability observed in conventional Image Restoration\(\rightarrow\)Object Detection cascade framework, we extend our analysis to the parameter space of the networks, where Lipschitz continuity characterizes the sensitivity of a model’s output to changes in its parameters. Our findings reveal that image restoration networks maintain relatively low Lipschitz constants, resulting in smooth and stable optimization trajectories during training. In contrast, object detection networks exhibit substantially higher Lipschitz constants, leading to sharp gradient transitions and erratic convergence paths. This imbalance disrupts gradient flow, introduces mutual interference, and destabilizes joint optimization, further compounding the instability of traditional cascade frameworks.

Given the importance of network stability in adverse conditions, a key challenge lies in harmonizing image restoration and object detection to address the inherent differences in Lipschitz continuity.
 To address this, we propose Lipschitz-regularized object detection (LROD), a simple yet effective framework that integrates image restoration directly into the detector’s feature learning. Unlike conventional cascades, LROD harmonizes the Lipschitz continuity of both tasks during training, smoothing out perturbations before they propagate through the detector’s discontinuous layers. This coupling mitigates noise amplification, enhancing stability in challenging environments. Furthermore, LROD introduces a parameter-space regularization term to stabilize gradient flows, ensuring smoother optimization dynamics and improved robustness under varying degradation intensities.

We implement LROD into existing YOLO detectors, taking advantage of their real-time performance, resource efficiency, and suitability for edge deployment. This integration yields an efficient model, called Lipschitz-regularized YOLO (LR-YOLO), which can be seamlessly applied to YOLO series detectors (\textit{e.g.}, YOLOv10~\cite{wang2024yolov10} and YOLOv8~\cite{yolov8}). 
As shown in Figure~\ref{fig:motivation}~(d), our Lipschitz-regularized object detection achieves a smoother Lipschitz continuity compared to the cascade framework.
Extensive experiments on image dehazing and low-light enhancement benchmarks demonstrate that LR-YOLO improves detection stability and robustness compared to traditional cascade frameworks.
In summary, our contributions are as follows:

\begin{itemize}[leftmargin=5pt]
\item 
\textbf{Lipschitz Continuity Analysis}: we perform a detailed analysis of Lipschitz continuity in both the input space and the parameter space of image restoration and object detection networks. Our analysis uncovers a critical mismatch in smoothness between these tasks, which potentially introduces instability and impedes effective integration.
{To our knowledge, this is the early work to provide a detailed Lipschitz continuity analysis aimed at understanding the instability challenges in cascade-based detection pipelines. }
\item 
\textbf{Lipschitz-Regularized Framework}: motivated by our analysis, we propose a simple and effective object detection framework that integrates image restoration directly into the detector’s feature learning, harmonizing the Lipschitz continuity of both tasks during training. This design enhances smoothness and mitigates the instability inherent in traditional cascade-based methods. 
\end{itemize}

\section{Related Work} 

\textbf{Object Detection Under Adverse Conditions.} 
Existing research primarily focuses on cascade frameworks, where image restoration techniques such as image dehazing~\cite{cui2023selective,cui2024revitalizing}, low-light enhancement~\cite{wang2023ultra,cai2023retinexformer}, and all-in-one restoration~\cite{conde2024instructir} are used as pre-processing steps to improve image quality and enhance human trust in detection results compared to domain adaptation approaches~\cite{zhao2024revisiting}. 
ReForDe~\cite{sun2022rethinking} uses adversarial training to generate detection-friendly labels for fine-tuning restoration networks.
SR4IR~\cite{kim2024beyond} introduces a training framework where image restoration is constrained by object detection, and detection training utilizes restoration outputs
Image-adaptive techniques~\cite{liu2022image,kalwar2023gdip} integrate differentiable image processing filters into the detection pipeline.
FeatEnHancer~\cite{hashmi2023featenhancer} applies hierarchical feature enhancement to improve detection performance.
Despite these advancements, the functional mismatch between restoration and detection networks is underexplored. Our work reports that the large disparity in Lipschitz continuity between the two tasks exacerbates non-smoothness when they are cascaded, leading to instability under varying degradation intensities. To address this, we propose a Lipschitz-regularized framework that enhances the Lipschitz continuity of the detection network, facilitating better harmonization between these two tasks.

\textbf{Lipschitz Continuity Analysis.} Lipschitz continuity is useful in analyzing the stability and robustness of deep neural networks~\cite{meunier2022dynamical,araujo2023unified,wang2023direct}. Models with lower Lipschitz constants tend to exhibit better generalization performance, especially under adversarial conditions~\cite{arora2018stronger}. This has motivated further research on regularization techniques that constrain the Lipschitz constant to enhance model robustness. For instance, SN-GAN~\cite{miyato2018spectral} controls the Lipschitz constant by restricting the spectral norm of network parameters, while other Lipschitz-based regularization techniques have been proposed to improve model stability~\cite{leino2021globally}. Several studies have extended these ideas to network design~\cite{qi2023lipsformer}, highlighting the critical role of Lipschitz continuity in controlling the smoothness and stability of neural networks. 
In our work, we analyze object detection stability under adverse conditions from both the input and parameter spaces using Lipschitz continuity as the lens of investigation. We demonstrate that the disparity in Lipschitz continuity between image restoration and object detection networks is a primary source of non-smoothness and instability in cascade frameworks.

\section{Lipschitz Continuity Perspective}
\label{sec:lipschitz}

\subsection{Input Space Analysis: Model Stability in Adverse Conditions}
\label{sec:model_stability}

\begin{figure}[t!]
    \centering
    \includegraphics[width=\linewidth]{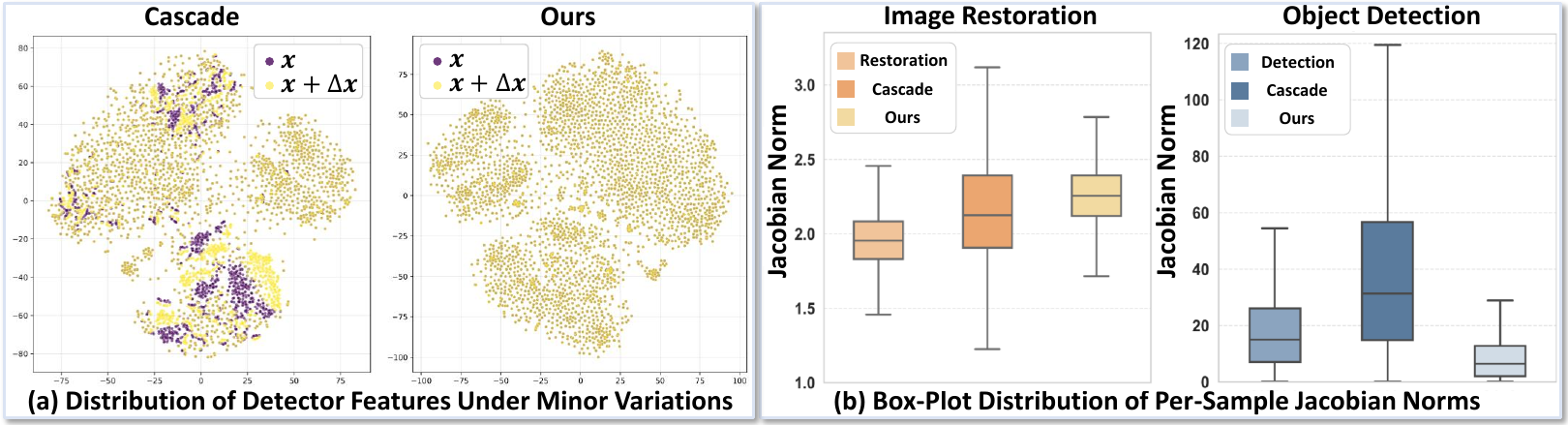}
    \caption{\textbf{Impact of haze density variations on feature stability and Lipschitz continuity.}
    (a) Distribution of the detector backbone's features between two haze inputs \(\boldsymbol{x}\) and \(\boldsymbol{x}+\Delta \boldsymbol{x}\) under minor haze density variations \(\Delta \boldsymbol{x}\) on Pascal VOC~\cite{everingham2015pascal} with synthetic haze. In the cascade framework, nearly half of the features shift under slight haze density variations, while our Lipschitz-regularized object detection remains stable. (b) Box-plot distribution of Jacobian norms \(\|\nabla_{\boldsymbol{x}} f_{\mathbf{\theta}} (\boldsymbol{x})\|\) at each sample \(\boldsymbol{x}\) 
    between image restoration and object detection task on Pascal VOC~\cite{everingham2015pascal} with synthetic haze. The Lipschitz constant of the object detection network is nearly an order of magnitude larger than that of the restoration network. This large disparity in Lipschitz continuity between the two tasks exacerbates the non-smoothness in the cascade framework. Our method constrains the Lipschitz constant of object detection to harmonize these two tasks better. ConvIR~\cite{cui2024revitalizing} and YOLOv8~\cite{yolov8} are taken as restoration and detection methods, respectively.}
    \label{fig:lipschitz}
\end{figure}

Object detection in adverse conditions, such as haze or low light, is highly sensitive to variations in degradation intensity, including changes in haze density and luminance fluctuations. Traditional Image Restoration\(\rightarrow\)Object Detection cascade framework struggles with such variations, leading to unstable detection results. As shown in Figure~\ref{fig:lipschitz}~(a), even when partially mitigated by restoration, minor perturbations still cause significant shifts in detector features, exposing the framework’s instability. To understand this, we analyze the problem through the lens of Lipschitz continuity, which quantifies a model’s sensitivity to input changes. Our findings reveal that the Lipschitz constant of the detection network is nearly an order of magnitude larger than that of the restoration network, amplifying noise and disrupting stability under adverse conditions.

We begin by recalling the definition of Lipschitz continuity: A network \( f(\cdot; \mathbf{\theta}): \mathbb{R}^D\mapsto\mathbb{R}^K \), defined on some domain \(dom(f)\subseteq\mathbb{R}^D\) with parameters \( \mathbf{\theta} \), is called \( C \)-Lipschitz continuous if there exists a real constant \(C>0\) such that \(\forall \boldsymbol{x}_1, \boldsymbol{x}_2 \in \operatorname{dom}(f): \|f(\boldsymbol{x}_1; \mathbf{\theta})-f(\boldsymbol{x}_2; \mathbf{\theta})\|_p\leq C\|\boldsymbol{x}_1-\boldsymbol{x}_2\|_p.\) For simplicity, we will compute the $2$-norm, denoted as \(\|\cdot\|\), throughout the rest of the paper, which can be easily generalized to other norms. Using Theorem 1 in~\cite{latorre2020lipschitz}, we know that for a differentiable, $C$-Lipschitz continuous network \( f(\cdot;\mathbf{\theta}): \mathbb{R}^D\mapsto\mathbb{R}^K \), the Lipschitz constant of \( f(\cdot;\mathbf{\theta}) \) can be expressed as \(C_{\boldsymbol{x}} (f(\boldsymbol{x}; \mathbf{\theta})) = \sup_{\boldsymbol{x}\in dom(f)} \|\nabla_{\boldsymbol{x}}f(\boldsymbol{x}; \mathbf{\theta})\|_* = \sup_{\boldsymbol{x}\in dom(f)} \|\nabla_{\boldsymbol{x}}f(\boldsymbol{x}; \mathbf{\theta})\|\), where \( \nabla_{\boldsymbol{x}}f(\boldsymbol{x}; \mathbf{\theta}) \) is Jacobian of \( f \) w.r.t. input \( \boldsymbol{x} \) and \(\|\cdot\|_*\) denotes the dual norm (The dual norm of the $2$-norm is itself).

To quantitatively assess the Lipschitz constant of the image restoration and object detection network, we compute the above Jacobian norm for each sample \(\boldsymbol{x}\) in the Pascal VOC dataset~\cite{everingham2015pascal}, considering variations in haze density. As shown in Figure~\ref{fig:lipschitz} (b), we observe that the Jacobian norm of the image restoration network ranges from $1$ to $3.5$ per sample, while the Jacobian norm of the object detection network is nearly an order of magnitude larger than that of the image restoration network. This indicates that object detection has a higher Lipschitz constant compared to image restoration. Therefore, the large disparity in Lipschitz continuity between the two tasks leads to an unstable framework when they are directly cascaded. Specifically, even slight variations will inevitably be amplified by the restoration network since its Jacobian norm per sample exceeds $1$, and further destabilized by the high-Lipschitz constant of the detection network within the cascade framework.

\begin{remark}
Image restoration networks exhibit smooth and continuous mappings, while object detection networks are more non-smooth from the perspective of Lipschitz continuity. This large disparity in Lipschitz continuity between the two tasks exacerbates the non-smoothness when they are directly cascaded, leading to instability under variations in degradation intensities.
\end{remark}

\subsection{Parameter Space Analysis: Training Stability}
\label{sec:trainging_stability}

The disparity in Lipschitz continuity between restoration and detection networks extends beyond the input space, impacting their training stability. To understand this, we analyze the parameter space of the networks
to capture how gradient updates influence model stability during optimization. Our analysis shows that restoration networks, with lower Lipschitz constants, maintain smooth optimization trajectories, while detection networks, with substantially higher Lipschitz constants, experience sharp gradient transitions and unstable convergence. This imbalance disrupts gradient flow, contributing to training instability in cascade-based designs.

We extend the Lipschitz continuity analysis to the parameter space: A network \( f(\boldsymbol{x};\mathbf{\theta}) \) defined on some parameter space \(\mathbf{\Theta}\) is called Lipschitz continuous in the parameter space if there exists \(C_{\mathbf{\theta}}(f(\boldsymbol{x};\mathbf{\theta})) > 0\) such that \(\forall \mathbf{\theta}_1, \mathbf{\theta}_2 \in \mathbf{\Theta} \), \( \|f(\boldsymbol{x};\mathbf{\theta}_1)-f(\boldsymbol{x};\mathbf{\theta}_2)\|\leq C_{\mathbf{\theta}}(f(\boldsymbol{x};\mathbf{\theta})) \|\mathbf{\theta}_1-\mathbf{\theta}_2\|. \) Due to the symmetry between \(\boldsymbol{x}\) and \(\mathbf{\theta}\), an analogous result holds when the two variables are interchanged: The Lipschitz constant in the parameter space of \( f(\boldsymbol{x};\mathbf{\theta}) \), defined on the parameter space \(\mathbf{\Theta}\), can be expressed as \( C_{\mathbf{\theta}}(f(\boldsymbol{x};\mathbf{\theta})) = \sup_{\mathbf{\theta}\in\mathbf{\Theta}} \|\nabla_{\mathbf{\theta}}f(\boldsymbol{x};\mathbf{\theta})\| \), where \( \nabla_{\mathbf{\theta}}f(\boldsymbol{x};\mathbf{\theta}) \) represents the gradient of network parameters in the parameter space.

Given that the network is trained using the gradient descent optimization algorithm, expressed as \( \mathbf{\theta} \leftarrow \mathbf{\theta} - \mu \cdot \nabla_{\mathbf{\theta}}f(\boldsymbol{x};\mathbf{\theta}) \) (\(\mu\) denotes the learning rate), the Lipschitz constant in the parameter space is crucial for ensuring training stability. This is because the Lipschitz constant in the parameter space acts as an upper bound for the gradients of the network parameters during training.

The Lipschitz continuity in parameter space reflects the sensitivity of the model’s output to variations in its parameters. To illustrate this, we visualize the loss landscape by perturbing parameters along two directions, revealing their impact on optimization smoothness and stability.
Specifically, we use the visualization method in~\cite{li2018visualizing}: let~\( \mathbf{\theta} \) represent the fixed model parameters, we select two normalized direction vectors \(\mathbf{\delta}\) and \(\mathbf{\eta}\) in the parameter space, and plot the function \(f(\alpha, \beta) = \mathcal{L}({\mathbf{\theta}}+\alpha \delta + \mathbf{\beta} \eta)\) on the surface, where \(\mathcal{L}\) is the loss function, and \(\alpha\) and \(\beta\) are the coordinates on the surface.

\begin{figure}[t!]
    \centering
    \includegraphics[width=\linewidth]{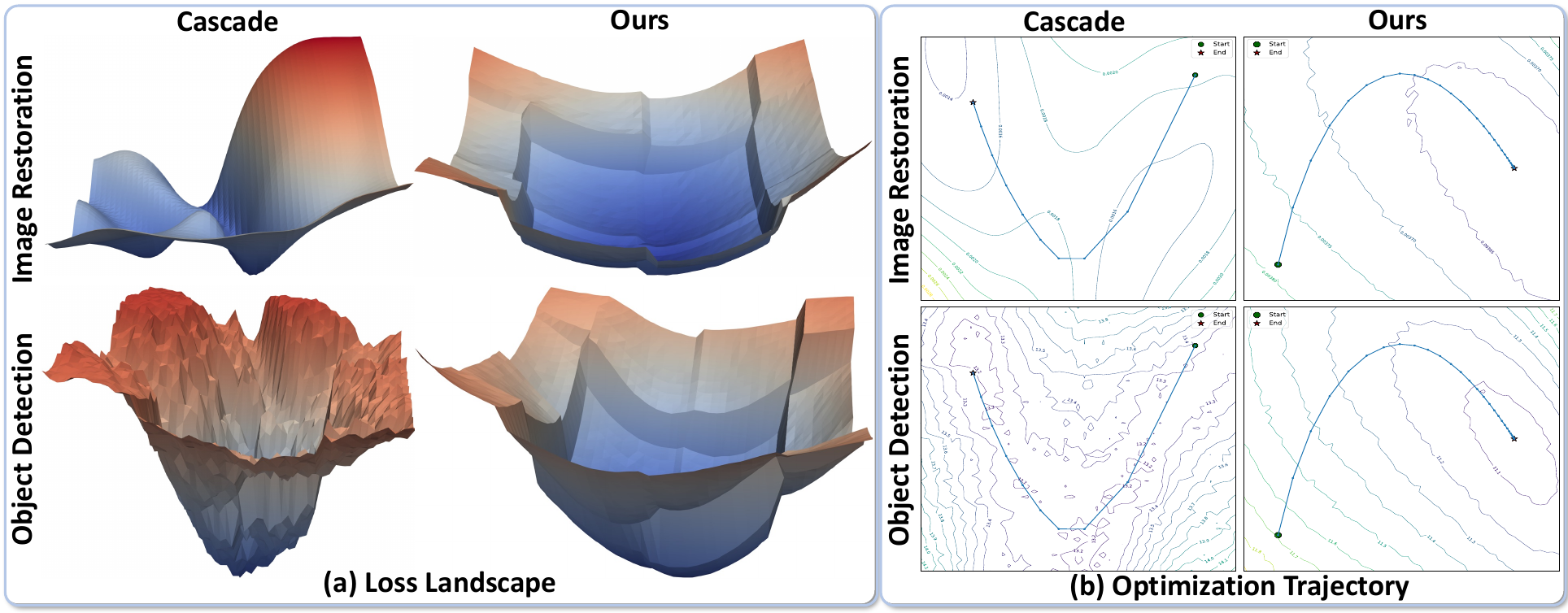}
    \caption{\textbf{Parameter-space smoothness and optimization stability comparison between the cascade framework and our Lipschitz-regularized object detection.}
    (a) Loss landscapes of restoration and detection tasks: restoration networks demonstrate smooth trajectories, while detection networks encounter sharp gradient transitions, indicating instability.
(b) The cascade framework amplifies this imbalance, leading to inefficient convergence and oscillatory optimization paths.
(c) Our method introduces Lipschitz regularization to smooth the parameter space of object detection, enhancing stability and harmonizing it with restoration. ConvIR~\cite{cui2024revitalizing} and YOLOv8~\cite{yolov8} are used as representative restoration and detection methods, respectively.}
    \label{fig:landscape}
\end{figure}

As shown in Figure~\ref{fig:landscape}~(a), the loss landscape of the image restoration network exhibits a smooth loss function, while the loss landscape of the object detection network is notably rough. 
This reflects differences in their Lipschitz constants in parameter space: restoration networks tend to have lower Lipschitz constants, giving smoother gradients, while detection networks exhibit higher Lipschitz constants, showing sharper transitions, and increased sensitivity to parameter changes.
Figure~\ref{fig:landscape} (b) further illustrates the optimization trajectories, where restoration follows stable paths, while detection experiences frequent shifts, indicating instability. This imbalance disrupts gradient flow during joint training, resulting in unstable convergence and reduced optimization efficiency.

\begin{remark}
Image restoration networks with lower Lipschitz constants exhibit smooth optimization trajectories, while object detection networks with higher Lipschitz constants experience sharp gradient transitions and unstable convergence. This imbalance in the parameter space between these two tasks results in training instability and reduced optimization efficiency in cascade-based designs.
\end{remark}

\section{Lipschitz-Regularized Object Detection}

The analysis in Section~\ref{sec:lipschitz} reveals the disparities in Lipschitz continuity between restoration and detection networks, manifesting in both the input space and the parameter space. Driven by this, we propose lipschitz-regularized object detection (LROD), a simple and effective framework that harmonizes restoration and detection through targeted Lipschitz regularization. Specifically, LCOD introduces two core mechanisms: \emph{1) Lipschitz regularization via low-Lipschitz restoration} to constrain the Lipschitz constant of object detection in the input space, and \emph{2) Lipschitz regularization via parameter-space smoothing} to constrain the Lipschitz constant in the parameter space. 

\subsection{Lipschitz Regularization via Low-Lipschitz Restoration}

Lipschitz continuity analysis in input space (Section~\ref{sec:model_stability}) shows that image restoration networks exhibit smooth, continuous mappings, while object detection networks are more non-smooth. 
By leveraging the low-Lipschitz properties of the restoration task, we integrate restoration learning into the detector backbone’s feature learning, constraining the Lipschitz constant of the object detection task in the input space. This better harmonizes the detection task with the low-Lipschitz restoration.

\begin{remark}[Lipschitz Regularization via Low-Lipschitz Restoration]
\label{remark:restoration}

Let:
\(
f_{\mathbf{\theta}_b, \mathbf{\theta}_d} = f_{\mathbf{\theta}_d} \circ f_{\mathbf{\theta}_b}
\)
denote the object detection model, where
\( f_{\mathbf{\theta}_b}(\cdot; \mathbf{\theta}_b) \) is the backbone network parameterized by \( \mathbf{\theta}_b \), and  
\( f_{\mathbf{\theta}_d}(\cdot; \mathbf{\theta}_d) \) is the detection head parameterized by \( \mathbf{\theta}_d \).
Similarly, let:
\(
\raisebox{0.3ex}{$g$}_{\mathbf{\theta}_b, \mathbf{\theta}_r} = f_{\mathbf{\theta}_r} \circ f_{\mathbf{\theta}_b}
\)
represent the image restoration model, where
\( f_{\mathbf{\theta}_r}(\cdot; \mathbf{\theta}_r) \) is the restoration head parameterized by \( \mathbf{\theta}_r \), sharing the same backbone \( f_{\mathbf{\theta}_b} \).
Given a weighted combination of the detection loss and the restoration loss:
\[
\mathcal{L}(\mathbf{\theta}_b, \mathbf{\theta}_d, \mathbf{\theta}_r) = \mathcal{L}_{\mathrm{det}}(f_{\mathbf{\theta}_b, \mathbf{\theta}_d}) + \lambda \cdot \mathcal{L}_{\mathrm{res}}(\raisebox{0.3ex}{$g$}_{\mathbf{\theta}_b, \mathbf{\theta}_r}), \quad \lambda > 0
\]

Let $\operatorname{Lip}(f_{\mathbf{\theta}_b}) := \sup_{\boldsymbol{x}} \| J_{f_{\mathbf{\theta}_b}}(\boldsymbol{x}) \|$ be the Lipschitz constant of $f_{\mathbf{\theta}_b}$ defined by jacobian norm. If: 
\begin{enumerate}[left=0pt, nosep]
    \item $\mathcal{L}_{\mathrm{res}}$ is Lipschitz continuous and $\|\nabla_{\mathbf{\theta}_b} \mathcal{L}_{\mathrm{res}}(\raisebox{0.3ex}{$g$}_{\mathbf{\theta}_b, \mathbf{\theta}_r})\| \le G$ for $G < \|\nabla_{\mathbf{\theta}_b} \mathcal{L}_{\mathrm{det}}(f_{\mathbf{\theta}_b, \mathbf{\theta}_d})\|$;
    \item There exists a training sample $\boldsymbol{x}^\star$ and $\gamma > 0$ such that: \(\left\langle \nabla_{\mathbf{\theta}_b} \| J_{f_{\mathbf{\theta}_b}}(\boldsymbol{x}^\star) \|, \nabla_{\mathbf{\theta}_b} \mathcal{L}_{\mathrm{res}}(\raisebox{0.3ex}{$g$}_{\mathbf{\theta}_b, \mathbf{\theta}_r}) \right\rangle \ge \gamma, \)
\end{enumerate}

then under continuous-time gradient descent \({\mathbf{\theta}_b}^{(t+1)} \leftarrow {\mathbf{\theta}_b}^{(t)} - \mu \cdot \nabla_{{\mathbf{\theta}_b}} \mathcal{L}(\mathbf{\theta}_b, \mathbf{\theta}_d, \mathbf{\theta}_r) \) (\(\mu\) denotes the learning rate), the evolution of the Lipschitz constant satisfies:
\[
\frac{d}{dt} \left[\operatorname{Lip}(f_{\mathbf{\theta}_b})\right] \le -\lambda \cdot \gamma + \xi(t)
\]
where $\xi(t) := \left\langle \nabla_{\mathbf{\theta}_b} \| J_{f_{\mathbf{\theta}_b}}(\boldsymbol{x}^\star) \|, \nabla_{\mathbf{\theta}_b} \mathcal{L}_{\mathrm{det}}(f_{\mathbf{\theta}_b, \mathbf{\theta}_d}) \right\rangle$ is the unconstrained change induced by the detection loss and \(\gamma\) is the regularization via the restoration task. 

This suggests that integrating the image restoration task directly into the detector’s feature learning by sharing the detector's backbone helps suppress the model’s sensitivity to input perturbations during training, effectively acting as a Lipschitz regularization.
The detailed proof is in Appendix~\ref{appendix:proof}.
\end{remark}

Specifically, we extract low-level features from the first three stages of the detector backbone, which preserve essential spatial and textural information for image restoration. These features are then passed through a restoration-specific head to obtain the restored images. By leveraging the inherently smoother Lipschitz continuity of the image restoration task, this restoration loss implicitly regularizes the feature representations used for object detection during training, thereby constraining the Lipschitz constant of the detection network in the input space. As shown in Figure~\ref{fig:lipschitz} (a) and (b), our Lipschitz-regularized object detection exhibits smoother Lipschitz continuity compared to both the original object detection and the cascade framework, with lower Lipschitz constants and more stable detector features under varying degradation intensities.

\subsection{Lipschitz Regularization via Parameter-Space Smoothing}

Lipschitz continuity analysis in parameter space (Section~\ref{sec:trainging_stability}) shows that low-Lipschitz restoration networks maintain smooth optimization trajectories, while high-Lipschitz detection networks experience sharp gradient transitions and unstable convergence. To improve harmony between these tasks and ensure training stability, we constrain the Lipschitz constant of the detection networks in the parameter space. We introduce a parameter-space regularization term to stabilize gradient flows, promoting smoother optimization dynamics.

\begin{remark}[Lipschitz Regularization via Parameter-Space Smoothing]
\label{remark:smooth}
Let $\mathbf{\theta} = \mathbf{\theta}_b \cup \mathbf{\theta}_d$ is the full parameter set of the detection model.
The parameter-space regularization term is defined as the gradient norm with respect to the model parameters, denoted by \(\left\| \nabla_{\mathbf{\theta}} f_{\mathbf{\theta}}(\boldsymbol{x}) \right\|\).

\end{remark}

\paragraph{Full framework.} The Lipschitz-regularized object detection (LROD) framework incorporates the above two regularizations to ensure stable and efficient training. The total loss function is defined as:
\[
\mathcal{L}_{\mathrm{total}} = \mathcal{L}_{\mathrm{det}} + \lambda \cdot \mathcal{L}_{\mathrm{res}} + \lambda_p \cdot \left\| \nabla_{\mathbf{\theta}} f_{\mathbf{\theta}}(\boldsymbol{x}) \right\|,
\]
where \(\mathcal{L}_{\mathrm{det}}\) is the detection loss, \(\mathcal{L}_{\mathrm{res}}\) is the restoration loss, computed as a Charbonnier loss~\cite{bruhn2005lucas} between the restored image and the ground truth clean image, and \(\left\| \nabla_{\mathbf{\theta}} f_{\mathbf{\theta}}(\boldsymbol{x}) \right\|\) is the regularization term. The weights \(\lambda\) and \(\lambda_p\) are used to balance the restoration and regularization terms, respectively. 

We implement this framework as Lipschitz-regularized YOLO (LR-YOLO), which builds upon YOLO detectors.
As illustrated in Figure~\ref{fig:landscape}, ConvIR~\cite{cui2024revitalizing} and YOLOv8~\cite{yolov8} are employed as representative restoration and detection methods, respectively. LR-YOLO smooths the loss landscape of object detection compared to traditional cascade frameworks, better aligning with image restoration during training. This results in smooth gradient flow, improved stability, and more efficient optimization.

\section{Experiments}

\subsection{Experimental Settings}

\textbf{Dataset.} Datasets cover two challenging conditions: \emph{hazy weather} and \emph{low-light environments}. 
For both settings, we use Pascal VOC~\cite{everingham2015pascal} and COCO~\cite{lin2014microsoft} datasets for training and validation following the degradation setting from~\cite{liu2022image,sun2022rethinking,wu2024unsupervised}, and real-world datasets for out-of-domain evaluation.

\emph{1) Training and Validation Data:} 
\emph{a) VOC\_Haze\_Train} and \emph{VOC\_Haze\_Val} consist of $8,111$ and $2,734$ images respectively. Haze is synthesized \emph{online} during training using the atmospheric scattering model with \(\beta \in [0.5, 1.5]\), while validation images are synthesized \emph{offline} once for reproducibility; 
\emph{b) VOC\_Dark\_Train} and \emph{VOC\_Dark\_Val} consist of $12,334$ and $3,760$ images respectively. Low-light is simulated \emph{online} during training and \emph{offline} for validation via gamma correction with \(\gamma \in [1.5, 5]\). 
The classes in both the training and validation datasets for haze and low-light conditions align with those in the real-world datasets;
\emph{c) COCO\_Haze\_Train} and \emph{COCO\_Dark\_Train} consist of $118,287$ training images, and \emph{COCO\_Haze\_Val} and \emph{COCO\_Dark\_Val} contain $5,000$ validation images.

\emph{2) Real-world Test Data.} We adopt \textbf{two} benchmark datasets for the out-of-domain evaluation: \emph{a) RTTS}~\cite{li2018benchmarking} contains $4,322$ real-world hazy images annotated with $5$ object categories, \emph{i.e.}, \emph{Person}, \emph{Car}, \emph{Bus}, \emph{Bicycle}, and \emph{Motorbike}; 
\emph{b) ExDark}~\cite{loh2019getting} contains $2,563$ real-world low-light images labeled with $10$ categories, \emph{i.e.}, \emph{People}, \emph{Car}, \emph{Bus}, \emph{Bicycle}, \emph{Motorbike}, \emph{Boat}, \emph{Bottle}, \emph{Chair}, \emph{Dog}, and \emph{Cat}.

\textbf{Evaluation Metrics.} We evaluate object detection performance using mean Average Precision (mAP) at an IoU threshold of \(50\%\), which excludes difficult objects by default. Additionally, we report \textbf{mAP$_{\text{difficult}}$}, which includes all objects, including challenging cases (\emph{e.g.}, occluded targets) on the Pascal VOC~\cite{everingham2015pascal} and RTTS~\cite{li2018benchmarking} datasets.
For the COCO dataset~\cite{lin2014microsoft}, we adopt standard COCO-style metrics, including mAP averaged over IoU thresholds from $0.5$ to $0.95$ (in $0.05$ increments), along with AP$_{50}$, AP$_{75}$, and scale-specific scores: AP$_S$ (small), AP$_M$ (medium), and AP$_L$ (large).

\textbf{Implementation Details:} 
We adopt YOLOv10-s and YOLOv8-s as the baseline detectors. For training, the loss weights are set to \(\lambda=10\) and \(\lambda_p=0.01\). We use the SGD optimizer with an initial learning rate of \(1\times10^{2}\) and a weight decay of \(5\times10^{-4}\). The model is trained on an RTX 4090 GPU for $100$ epochs with a batch size of $16$, requiring approximately $8$ hours.
Input images are resized to \(640\times640\), and standard YOLO data augmentation techniques (\emph{e.g.}, random flipping and affine transformation) are applied. 
For experiments on the COCO dataset, we use $8$ RTX 4090 GPUs with a batch size of $16$ per GPU. Training is conducted for $300$ epochs and takes approximately $48$ hours.

\subsection{Object Detection under Adverse Conditions}

\begin{table*}[!t]
\centering
\caption{
\textbf{Comparison under two adverse conditions: \emph{haze weather} and \emph{low-light environment}.} Left: Results on \emph{VOC\_Haze\_Val} and \emph{RTTS}~\cite{li2018benchmarking}, with models trained on \emph{VOC\_Haze\_Train}. Right: Results on \emph{VOC\_Dark\_Val} and \emph{ExDark}~\cite{loh2019getting}, with models trained on \emph{VOC\_Dark\_Train}. In the cascade framework, $\dagger$ indicates adversarial training~\cite{sun2022rethinking}, and $\ddagger$ denotes alternating training~\cite{kim2024beyond}.
}
\label{tab:weather_det}
\tiny
\begin{minipage}[t]{0.50\textwidth}
\centering
\resizebox{\textwidth}{!}{
\begin{tabular}{l|>{\columncolor{mygray}}cc|>{\columncolor{mygray}}cc}
\hline
\multirow{3}{*}{\textbf{Methods}} & \multicolumn{4}{c}{\cellcolor{myblue}\textbf{\emph{Datasets (Haze Weather)}}} \\ 
 & \multicolumn{2}{c|}{\textbf{\emph{VOC\_Haze\_Val}}} & \multicolumn{2}{c}{\textbf{\emph{RTTS}}~\cite{li2018benchmarking}} \\ 
& \textbf{mAP} & \textbf{mAP$_{\text{difficult}}$} & \textbf{mAP} & \textbf{mAP$_{\text{difficult}}$} \\
\hline
YOLOv10~\cite{wang2024yolov10} & 50.5 & 44.7 & 42.6 & 33.8 \\
SFNet~\cite{cui2023selective}$\rightarrow$YOLOv10 & 77.9 & 70.1 & 45.5 & 35.9 \\
SFNet~\cite{cui2023selective}$\rightarrow$YOLOv10$^\dagger$~\cite{sun2022rethinking} & 79.1 & 72.1 & 46.6 & 37.1 \\
SFNet~\cite{cui2023selective}$\rightarrow$YOLOv10$^\ddagger$~\cite{kim2024beyond} & 79.3 & 71.7 & 45.8 & 36.0 \\
ConvIR~\cite{cui2024revitalizing}$\rightarrow$YOLOv10 & 79.9 & 72.2 & 46.1 & 36.0 \\
ConvIR~\cite{cui2024revitalizing}$\rightarrow$YOLOv10$^\dagger$~\cite{sun2022rethinking} & 80.1 & \underline{72.9} & 46.6 & \underline{37.2} \\
ConvIR~\cite{cui2024revitalizing}$\rightarrow$YOLOv10$^\ddagger$~\cite{kim2024beyond} & \underline{80.5} & 72.6 & 46.5 & 36.5 \\
IA~\cite{liu2022image}$\rightarrow$YOLOv10 & 79.9 & 72.0 & 45.4 & 35.8 \\
GDIP~\cite{kalwar2023gdip}$\rightarrow$YOLOv10 & 79.2 & 70.9 & \underline{47.2} & 37.0 \\
FeatEnHancer~\cite{hashmi2023featenhancer}$\rightarrow$YOLOv10 & 79.8 & 71.6 & 46.7 & 36.2 \\
\rowcolor{myblue}
LR-YOLOv10 (Ours) & \textbf{82.5} & \textbf{74.4} & \textbf{49.2} & \textbf{38.5} \\
\hline
YOLOv8~\cite{yolov8} & 54.3 & 48.3 & 45.3 & 36.2 \\
SFNet~\cite{cui2023selective}$\rightarrow$YOLOv8 & 79.2 & 71.1 & 48.9 & 38.4 \\
SFNet~\cite{cui2023selective}$\rightarrow$YOLOv8$^\dagger$~\cite{sun2022rethinking} & 80.8 & 73.8 & 49.1 & 39.3 \\
SFNet~\cite{cui2023selective}$\rightarrow$YOLOv8$^\ddagger$~\cite{kim2024beyond} & 80.3 & 72.8 & 49.3 & 39.2 \\
ConvIR~\cite{cui2024revitalizing}$\rightarrow$YOLOv8 & 80.5 & 72.8 & 49.3 & 38.7 \\
ConvIR~\cite{cui2024revitalizing}$\rightarrow$YOLOv8$^\dagger$~\cite{sun2022rethinking} & 80.9 & \underline{74.1} & 49.5 & 39.0 \\
ConvIR~\cite{cui2024revitalizing}$\rightarrow$YOLOv8$^\ddagger$~\cite{kim2024beyond} & \underline{81.4} & 74.0 & 50.1 & \underline{39.9} \\
IA~\cite{liu2022image}$\rightarrow$YOLOv8 & 80.6 & 73.0 & 47.7 & 37.3 \\
GDIP~\cite{kalwar2023gdip}$\rightarrow$YOLOv8 & 81.0 & 73.1 & \underline{50.3} & 39.8 \\
FeatEnHancer~\cite{hashmi2023featenhancer}$\rightarrow$YOLOv8 & 81.2 & 73.4 & 48.4 & 38.8 \\
\rowcolor{myblue}
LR-YOLOv8 (Ours) & \textbf{83.3} & \textbf{76.5} & \textbf{53.2} & \textbf{42.4} \\
\hline
\end{tabular}
}
\end{minipage}
\hfill
\begin{minipage}[t]{0.48\textwidth}
\centering
\resizebox{\textwidth}{!}{
\begin{tabular}{l|>{\columncolor{mygray}}cc|>{\columncolor{mygray}}c}
\hline
\multirow{3}{*}{\textbf{Methods}} & \multicolumn{3}{c}{\cellcolor{myblue}\textbf{\emph{Datasets (Low-Light Environment)}}} \\
 & \multicolumn{2}{c|}{\textbf{\emph{VOC\_Dark\_Val}}} & \multicolumn{1}{c}{\textbf{\emph{ExDark}}~\cite{loh2019getting}} \\
& \textbf{mAP} & \textbf{mAP$_{\text{difficult}}$} & \textbf{mAP} \\
\hline
YOLOv10~\cite{wang2024yolov10} & 62.1 & 55.0 & 49.2 \\
LLFormer~\cite{wang2023ultra}$\rightarrow$YOLOv10 & 65.6 & 58.0 & 46.3 \\
LLFormer~\cite{wang2023ultra}$\rightarrow$YOLOv10$^\dagger$~\cite{sun2022rethinking} & 64.7 & 57.5 & 47.0 \\
LLFormer~\cite{wang2023ultra}$\rightarrow$YOLOv10$^\ddagger$~\cite{kim2024beyond} & 66.3 & 59.2 & 49.5 \\
Retinexformer~\cite{cai2023retinexformer}$\rightarrow$YOLOv10 & 66.3 & 58.6 & 47.6 \\
Retinexformer~\cite{cai2023retinexformer}$\rightarrow$YOLOv10$^\dagger$~\cite{sun2022rethinking} & 66.0 & 58.4 & 45.8 \\
Retinexformer~\cite{cai2023retinexformer}$\rightarrow$YOLOv10$^\ddagger$~\cite{kim2024beyond} & 66.9 & 59.2 & 47.5 \\
IA~\cite{liu2022image}$\rightarrow$YOLOv10 & 66.0 & 58.7 & 50.4 \\
GDIP~\cite{kalwar2023gdip}$\rightarrow$YOLOv10 & 65.8 & 58.5 & 48.9 \\
FeatEnHancer~\cite{hashmi2023featenhancer}$\rightarrow$YOLOv10 & \underline{67.6} & \underline{59.9} & \underline{50.9} \\
\rowcolor{myblue}
LR-YOLOv10 (Ours) & \textbf{70.6} & \textbf{62.7} & \textbf{53.8} \\
\hline
YOLOv8~\cite{yolov8} & 63.4 & 55.8 & 50.0 \\
LLFormer~\cite{wang2023ultra}$\rightarrow$YOLOv8 & 66.2 & 58.7 & 46.6 \\
LLFormer~\cite{wang2023ultra}$\rightarrow$YOLOv8$^\dagger$~\cite{sun2022rethinking} & 66.2 & 58.8 & 47.9 \\
LLFormer~\cite{wang2023ultra}$\rightarrow$YOLOv8$^\ddagger$~\cite{kim2024beyond} & 66.2 & 59.2 & 48.6 \\
Retinexformer~\cite{cai2023retinexformer}$\rightarrow$YOLOv8 & 67.8 & 59.5 & 47.6 \\
Retinexformer~\cite{cai2023retinexformer}$\rightarrow$YOLOv8$^\dagger$~\cite{sun2022rethinking} & 67.7 & 60.0 & 49.5 \\
Retinexformer~\cite{cai2023retinexformer}$\rightarrow$YOLOv8$^\ddagger$~\cite{kim2024beyond} & 68.6 & 61.0 & 49.5 \\
IA~\cite{liu2022image}$\rightarrow$YOLOv8 & 66.5 & 59.2 & 49.6 \\
GDIP~\cite{kalwar2023gdip}$\rightarrow$YOLOv8 & \underline{68.9} & \underline{61.2} & 51.2 \\
FeatEnHancer~\cite{hashmi2023featenhancer}$\rightarrow$YOLOv8 & 68.7 & 60.8 & \underline{51.8} \\
\rowcolor{myblue}
LR-YOLOv8 (Ours) & \textbf{71.7} & \textbf{63.9} & \textbf{54.5} \\
\hline
\end{tabular}
}
\end{minipage}
\end{table*}

Table~\ref{tab:weather_det} presents a method comparison of object detection under two adverse conditions: \emph{hazy weather}, evaluated on the \emph{VOC\_Haze\_Val} and \emph{RTTS}~\cite{li2018benchmarking} datasets, and \emph{low-light environments}, evaluated on the \emph{VOC\_Dark\_Val} and \emph{ExDark}~\cite{loh2019getting} datasets. 
We compare various image restoration methods, including SFNet~\cite{cui2023selective}, ConvIR~\cite{cui2024revitalizing}, LLFormer~\cite{wang2023ultra}, and RetinexFormer~\cite{cai2023retinexformer}, all of which are trained on degraded images and used to restore inputs before detection. 
We further consider two joint training strategies: \emph{1) Adversarial training}~\cite{sun2022rethinking}, where restoration networks are fine-tuned to generate detection-friendly images; \emph{2) Alternating training}~\cite{kim2024beyond}, where restoration is supervised using detection-driven perceptual losses and detection is trained on restored outputs.
Furthermore, we include end-to-end methods for comparison, including IA~\cite{liu2022image}, GDIP~\cite{kalwar2023gdip}, and FeatEnHancer~\cite{hashmi2023featenhancer}. They are trained directly on degraded inputs. All models are trained from scratch on the \emph{VOC\_Haze\_Train} and \emph{VOC\_Dark\_Train} datasets, respectively. 
Our method outperforms other methods when using both YOLOv10 and YOLOv8 as object detectors, achieving mAP improvements of $2.0$ and $2.9$ on \emph{RTTS}, and $2.9$ and $2.7$ on \emph{ExDark}, respectively. 
Table~\ref{tab:coco_det} shows a comparison on \emph{COCO\_Haze\_Val} and \emph{COCO\_Dark\_Val} datasets, trained on \emph{COCO\_Haze\_Train} and \emph{COCO\_Dark\_Train}, respectively. We compare the all-in-one restoration 
method InstructIR~\cite{conde2024instructir}. Our method consistently improves performance across all evaluation metrics, achieving mAP improvements of $1.0$ and $1.2$, respectively.

\subsection{Evaluation and Analysis}

\begin{table}[!t]
\caption{
\textbf{Comparison on \emph{COCO\_Haze\_Val} and \emph{COCO\_Dark\_Val} datasets under haze and low-light conditions.} All models are trained from scratch on \emph{COCO\_Haze\_Train} and \emph{COCO\_Dark\_Train}, respectively. 
The $\dagger$ indicates adversarial training~\cite{sun2022rethinking} and $\ddagger$ denotes alternating training~\cite{kim2024beyond}.
}
\label{tab:coco_det}
\centering
\tiny
\resizebox{\textwidth}{!}{
\begin{tabular}{l|>{\columncolor{mygray}}cccccc|>{\columncolor{mygray}}cccccc}
\hline
\multirow{2}{*}{\textbf{Methods}} & \multicolumn{6}{c|}{\textbf{\emph{COCO\_Haze\_Val}}} & \multicolumn{6}{c}{\textbf{\emph{COCO\_Dark\_Val}}} \\ %
& \textbf{mAP} & \textbf{AP$_{50}$} & \textbf{AP$_{75}$} & \textbf{AP$_{S}$} & \textbf{AP$_{M}$} & \textbf{AP$_{L}$} & \textbf{mAP} & \textbf{AP$_{50}$} & \textbf{AP$_{75}$} & \textbf{AP$_{S}$} & \textbf{AP$_{M}$} & \textbf{AP$_{L}$} \\
\hline
YOLOv8~\cite{yolov8} & 20.3 & 28.8 & 22.0 & 9.1 & 22.8 & 29.1 & 31.3 & 45.2 & 33.5 & 16.5 & 34.2 & 45.1  \\
InstructIR~\cite{conde2024instructir}$\rightarrow$YOLOv8 & 33.8 & 47.7 & 36.6 & 15.6 & 37.2 & 50.3 & 31.1 & 44.8 & 33.3 & 15.0 & 33.7 & 46.5 \\
InstructIR~\cite{conde2024instructir}$\rightarrow$YOLOv8$^\dagger$~\cite{sun2022rethinking} & 33.4 & 47.7 & 36.1 & 15.2 & 36.8 & 49.0 & 30.2 & 43.8 & 32.3 & 14.3 & 32.9 & 44.5 \\
InstructIR~\cite{conde2024instructir}$\rightarrow$YOLOv8$^\ddagger$~\cite{kim2024beyond} & 35.0 & 49.7 & 37.9 & 17.0 & 38.9 & 51.1 & 30.4 & 43.8 & 32.7 & 15.5 & 32.9 & 45.2 \\
IA~\cite{liu2022image}$\rightarrow$YOLOv8 & 36.4 & 51.4 & 39.5 & 18.0 & 39.8 & 51.6 & 33.3 & 47.8 & 36.0 & 17.7 & 36.1 & 48.2   \\
GDIP~\cite{kalwar2023gdip}$\rightarrow$YOLOv8 & 36.6 & 51.8 & 39.4 & 18.1 & 40.6 & 51.5 & 33.4 & 47.6 & 36.0 & 17.5 & 35.9 & 47.6 \\
FeatEnHancer~\cite{hashmi2023featenhancer}$\rightarrow$YOLOv8 & \underline{36.7} & 52.2 & 39.7 & 18.3 & 40.5 & 52.1 & \underline{34.1} & 49.2 & 37.1 & 17.9 & 37.0 & 48.5 \\
\rowcolor{myblue}
LR-YOLOv8 (Ours) & \textbf{37.7} & 53.3 & 40.6 & 19.5 & 41.6 & 52.7 & \textbf{35.3} & 50.5 & 37.9 & 19.0 & 38.3 & 49.7 \\
\hline
\end{tabular}
}
\end{table}

\begin{table}[!t]
\centering
\tiny
\resizebox{\textwidth}{!}{
\begin{tabular}{c|cccccccc>{\columncolor{myblue}}c}
\hline
Methods & SFNet~\cite{cui2023selective} & ConvIR~\cite{cui2024revitalizing} & LLFormer~\cite{wang2023ultra} & Retinexformer~\cite{cai2023retinexformer} & InstructIR~\cite{conde2024instructir} & IA~\cite{liu2022image} & GDIP~\cite{kalwar2023gdip} & FeatEnHancer~\cite{hashmi2023featenhancer} & \textbf{Ours} \\
\hline
Params (M) & 13.27 & 5.53 & 24.55 & 1.61 & 31.15 & 0.17 & 138.24 & 0.14 & 0.52 \\
Flops (G) & 775.17 & 42.10 & 22.52 & 15.57 & 123.90 & 12.32 & 40.37 & 44.29 & 11.32 \\
\hline
\end{tabular}
}
\caption{
\textbf{Computational complexity comparison.}  Our method shows lower computational complexity in terms of the number of parameters (Params) and floating point operations (FLOPs).
}
\label{tab:flops}
\end{table}

\textbf{Computational Complexity Evaluation.} Table~\ref{tab:flops} presents a comparison of parameters (Params) and floating point operations (FLOPs), showcasing the inference efficiency of our framework.

\begin{figure}[t]
    \centering
    \begin{floatrow}
        \ffigbox[\FBwidth]{
            \begin{minipage}[t]{0.48\textwidth}
            \tiny
            \centering
            \resizebox{\textwidth}{!}{
            \begin{tabular}{c|cc|cc}
                \toprule
                & \(\mathcal{L}_{\mathrm{res}}\) & \(\| \nabla_{{\mathbf{\theta}}} f_{{\mathbf{\theta}}}(\boldsymbol{x}) \|\) & \textbf{\emph{RTTS}}~\cite{li2018benchmarking} & \textbf{\emph{ExDark}}~\cite{loh2019getting} \\
                \midrule
                \multirow{4}{*}{\rotatebox{90}{YOLOv10}} & & & 46.0 & 50.6 \\
                & \( \checkmark \) & & 48.1 & 52.7 \\
                & & \( \checkmark \) & 47.2 & 51.5 \\
                & \( \checkmark \) & \( \checkmark \) & \textbf{49.2} & \textbf{53.8} \\
                \midrule
                \multirow{4}{*}{\rotatebox{90}{YOLOv8}} & & & 49.3 & 51.6 \\
                & \( \checkmark \) & & 51.3 & 53.6 \\
                & & \( \checkmark \) & 50.1 & 52.4 \\
                & \( \checkmark \) & \( \checkmark \) & \textbf{53.2} & \textbf{54.5} \\
                \bottomrule
            \end{tabular}
            }
            \end{minipage}
        }{
            \captionof{table}{\textbf{Lipschitz regularization ablation study.} We evaluate the effect of two Lipschitz regularization parts \(\mathcal{L}_{\mathrm{res}}\) and \(\| \nabla_{{\mathbf{\theta}}} f_{{\mathbf{\theta}}}(\boldsymbol{x}) \|\).}
            \label{tab:ablation}
        }
        \ffigbox[\FBwidth]{
            \includegraphics[width=0.92\linewidth]{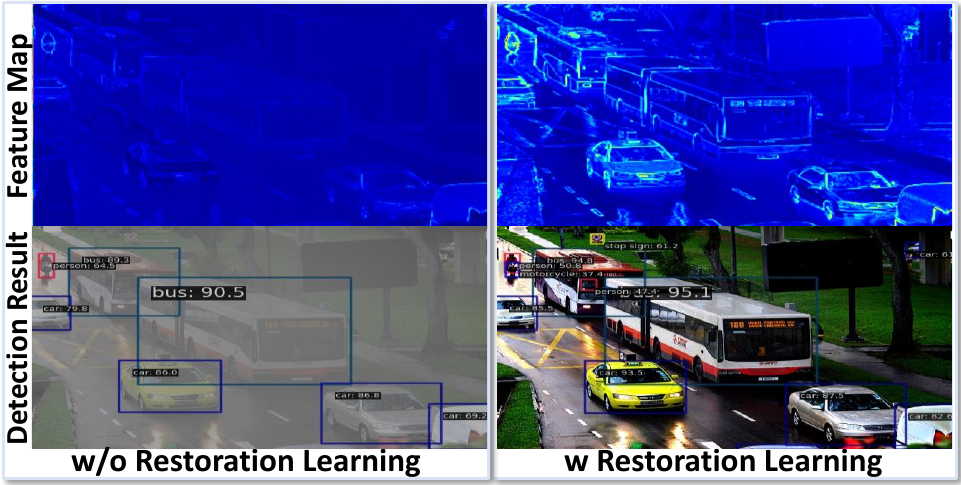}
        }{
            \caption{\textbf{Feature visualization.} We visualize the feature maps in the backbone of our model trained without and with \(\mathcal{L}_{\mathrm{res}}\).}
            \label{fig:feature}
        }

    \end{floatrow}
\end{figure}

\textbf{Lipschitz Regularization Ablation Study.} 
We evaluate the impact of two Lipschitz regularization parts (low-Lipschitz restoration learning \(\mathcal{L}_{\mathrm{res}}\) and parameter-space smoothing \(\| \nabla_{{\mathbf{\theta}}} f_{{\mathbf{\theta}}}(\boldsymbol{x}) \|\)). The evaluation is conducted on the \emph{RTTS} and \emph{ExDark} for out-of-domain performance, as presented in Table~\ref{tab:ablation}. 
Incorporating both restoration learning and parameter space smoothing during training improves synergy between detection and restoration, leading to mAP gains of $3.2$ for YOLOv10 and $3.9$ for YOLOv8 in \emph{RTTS}, and $3.2$ and $2.9$ on \emph{ExDark}, respectively, compared to baseline methods.

\textbf{Restoration Learning Analysis.} We visualize the backbone features of our model trained with and without restoration learning \(\mathcal{L}_{\mathrm{res}}\) as shown in Figure~\ref{fig:feature}. Integrating restoration learning into the detector’s feature learning facilitates the enhancement of degraded image features in the backbone, resulting in improved detection (\emph{e.g.}, complete detections of objects like the stop sign and motorcycle).

\begin{wraptable}{r}{0.38\linewidth}
\vspace{-\baselineskip}
\centering
\tiny
\setlength{\tabcolsep}{5pt}
\caption{\textbf{Alternative regularization ablation study.}}
\label{tab:alt-reg}
\begin{tabular}{lcccc}
\toprule
Method & Baseline & SNR~\cite{miyato2018spectral} & PGD~\cite{madry2017towards} & Ours \\
\midrule
\textit{RTTS} & 49.3 & 50.1 & 40.8 & \textbf{53.2} \\
\bottomrule
\end{tabular}
\vspace{-0.8\baselineskip}
\end{wraptable}
\textbf{Alternative Regularization Ablation Study.} 
We compare our method with two alternative regularization strategies—Spectral Norm Regularization (SNR~\cite{miyato2018spectral}) and adversarial training via PGD~\cite{madry2017towards}—by training on \textit{VOC\_Haze\_Train} and evaluating out-of-domain on \textit{RTTS}. As presented in Table~\ref{tab:alt-reg}, our approach attains the best \textit{RTTS} performance, surpassing both SNR and PGD. Unlike SNR, which constrains weights globally, our method penalizes \(\left\| \nabla_{\mathbf{\theta}} f_{\mathbf{\theta}}(\boldsymbol{x}) \right\|\), reducing output sensitivity to parameter changes and enabling input-aware smoothness. Compared to PGD-based adversarial training, which requires generating perturbed inputs and increases training cost, our approach achieves implicit robustness without adversarial examples, resulting in more stable and efficient training and no observed degradation on clean inputs.

\begin{wraptable}{r}{0.38\linewidth}
\vspace{-\baselineskip}
\centering
\tiny
\setlength{\tabcolsep}{4pt}
\caption{\textbf{Alternative variants ablation study.}}
\label{tab:share}
\begin{tabular}{lcccc}
\toprule
Sharing & Baseline & F1--F2 & F1--F3 (Ours) & F1--F4 \\
\midrule
\textit{RTTS} & 49.3 & 51.9 & \textbf{53.2} & 52.8 \\
\bottomrule
\end{tabular}
\vspace{-0.8\baselineskip}
\end{wraptable}
\textbf{Architectural Variants Ablation Study.} We ablate how deeply to share the encoder between detection and restoration by varying the number of shared stages, training on \textit{VOC\_Haze\_Train} and evaluating out-of-domain on \textit{RTTS}. As summarized in Table~\ref{tab:share}, shallower sharing (F1–F2) provides insufficient regularization, while deeper sharing (F1–F4) introduces task interference, supporting our design choice of the first three stages (F1–F3).

\begin{wraptable}{r}{0.38\linewidth}
\vspace{-\baselineskip}
\centering
\scriptsize
\setlength{\tabcolsep}{4pt}
\caption{\textbf{Regularization coefficients ablation study.}}
\label{tab:lambda-ablation}
\begin{tabular}{c|cccccc}
\toprule
\(\lambda\) & 0 & 10 & 10 & 10 & 20 & 5 \\
\(\lambda_p\) & 0 & 0.005 & 0.02 & 0.01 & 0.01 & 0.01 \\
\midrule
\textit{RTTS} & 49.3 & 52.9 & 53.0 & \textbf{53.2} & 53.1 & 52.8 \\
\bottomrule
\end{tabular}
\vspace{-0.8\baselineskip}
\end{wraptable}
\textbf{Regularization Coefficients Ablation Study.} We ablate the input-space and parameter-space regularization strengths by sweeping \( \lambda \in \{0,5,10,20\} \) and \( \lambda_p \in \{0,0.005,0.01,0.02\} \), training on \textit{VOC\_Haze\_Train} and evaluating out-of-domain on \textit{RTTS}. As reported in Table~\ref{tab:lambda-ablation}, our method consistently outperforms the baseline across a range of coefficient values, with only minor performance variation over the sweep, demonstrating robustness to the choice of regularization magnitudes.

\begin{figure}[t!]
    \centering
    \includegraphics[width=\linewidth]{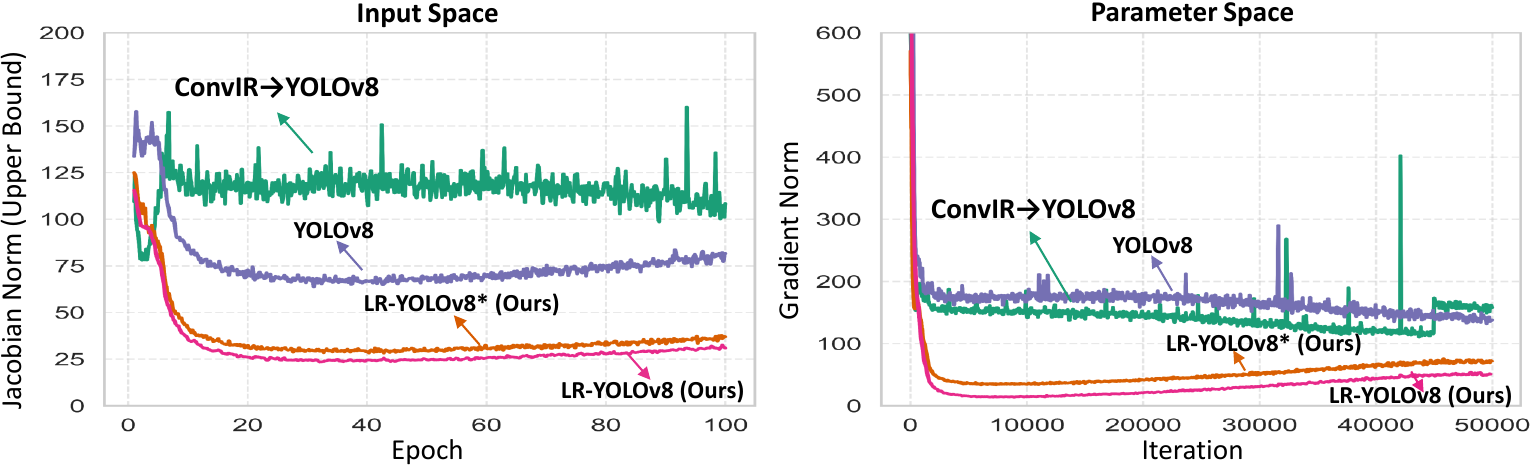}
    \caption{
    \textbf{Comparison of changes in Lipschitz continuity in both the input space and parameter space during training.} Methods include ConvIR$\rightarrow$YOLOv8 (Cascade), YOLOv8 (Baseline), LR-YOLOv8* (only trained with \(\mathcal{L}_{\mathrm{res}}\)), and LR-YOLOv8 (trained with both \(\mathcal{L}_{\mathrm{res}}\) and \(\| \nabla_{{\mathbf{\theta}}} f_{{\mathbf{\theta}}}(\boldsymbol{x}) \|\)). 
    }
    \label{fig:analysis}
\end{figure}

\textbf{Lipschitz Continuity Analysis.} We analyze changes in Lipschitz continuity in both the input and parameter spaces during training. Specifically, we monitor the upper bound of the Jacobian norm \(\sup_{\boldsymbol{x}\in dom(f)} \|\nabla_{\boldsymbol{x}}f_{\mathbf{\theta}}(\boldsymbol{x})\|\) in the input space, where \(dom(f)\) represents the domain of input images from Pascal VOC. Additionally, we track the gradient norm \(\|\nabla_{\mathbf{\theta}}f_{\mathbf{\theta}}(\boldsymbol{x})\|\) in the parameter space. As shown in Figure~\ref{fig:analysis}, LR-YOLOv8 trained with \(\mathcal{L}_{\mathrm{res}}\) reduces the Lipschitz constant in both the input and parameter spaces compared to ConvIR\(\rightarrow\)YOLOv8 and YOLOv8 during training. Training with~\(\|\nabla_{\mathbf{\theta}}f_{\mathbf{\theta}}(\boldsymbol{x})\|\) further promotes Lipschitz continuity in the input and parameter spaces.

\textbf{Generalization on Other Detection Paradigm.} We further validate the generalizability of our LROD by integrating it into the shared backbones of a transformer-based detector (RT-DETR~\cite{zhao2024detrs}) and a two-stage detector (Faster R-CNN~\cite{ren2015faster}). All models are trained on \textit{VOC\_Haze\_Train} and evaluated on both the synthetic \textit{VOC\_Haze\_Val} and real-world \textit{RTTS} dataset. As shown in Table~\ref{tab:main_rtdetr_frcnn}, LROD consistently outperforms other methods when using other detection paradigms, achieving mAP improvements of 1.4 on \textit{RTTS}, supporting LROD as a plug-and-play regularization framework.

\textbf{Generalization on Other Degradation.} We further assess robustness under additional adverse conditions—motion blur, rain, snow, and a haze–rain mixture—by constructing matched train/validation splits for each degradation and retraining all methods per setting. As reported in Table~\ref{tab:robustness_yolov8}, our method consistently outperforms existing methods, achieving mAP gains of 2.1, 2.6, 2.5, and 2.4, respectively. These results highlight the versatility and robustness of our method across diverse degradation.

\begin{figure}[t]
    \centering
    \begin{floatrow}
        \ffigbox[\FBwidth]{
            \begin{minipage}[t]{0.44\textwidth}
            \tiny
            \centering
            \setlength\tabcolsep{2pt}
            \resizebox{\textwidth}{!}{
            \begin{tabular}{lcccc}
            \toprule
            \multirow{2}{*}{Method} & \multicolumn{2}{c}{RT-DETR~\cite{zhao2024detrs}} & \multicolumn{2}{c}{Faster R-CNN~\cite{ren2015faster}} \\
            \cmidrule(lr){2-3} \cmidrule(lr){4-5}
             & \textit{VOC\_Haze\_Val} & \textit{RTTS} & \textit{VOC\_Haze\_Val} & \textit{RTTS} \\
            \midrule
            Baseline        & 51.5 & 43.7 & 69.1 & 43.2 \\
            ConvIR~\cite{cui2024revitalizing}          & 76.0 & 43.7 & 78.5 & 44.1 \\
            IA~\cite{liu2022image}              & 76.8 & 43.6 & 78.6 & 41.3 \\
            GDIP~\cite{kalwar2023gdip}            & 72.6 & 43.5 & 76.6 & 44.5 \\
            FeatEnHancer~\cite{hashmi2023featenhancer}    & 73.3 & 42.4 & 77.7 & 39.4 \\
            \rowcolor{myblue}
            LROD (Ours)     & \textbf{78.9} & \textbf{45.1} & \textbf{80.2} & \textbf{45.9} \\
            \bottomrule
            \end{tabular}
            }
            \end{minipage}
        }{
            \captionof{table}{\textbf{Generalization on other detection paradigm.} We integrate our LROD into RT-DETR~\cite{zhao2024detrs} and Faster-RCNN~\cite{ren2015faster}.}
            \label{tab:main_rtdetr_frcnn}
        }
        \ffigbox[\FBwidth]{
            \begin{minipage}[t]{0.48\textwidth}
            \tiny
            \centering
            \renewcommand\arraystretch{1.25}
            \setlength\tabcolsep{2pt}
            \resizebox{\textwidth}{!}{
            \begin{tabular}{lcccc}
            \toprule
            Method & Motion Blur & Rain & Snow & Haze + Rain \\
            \midrule
            YOLOv8~\cite{yolov8}                   & 50.8 & 53.1 & 60.8 & 50.1 \\
            ConvIR~\cite{cui2024revitalizing}$\rightarrow$YOLOv8~\cite{yolov8}     & 80.1 & 79.9 & 80.5 & 79.2 \\
            IA~\cite{liu2022image}$\rightarrow$YOLOv8~\cite{yolov8}         & 79.6 & 79.9 & 80.3 & 78.0 \\
            GDIP~\cite{kalwar2023gdip}$\rightarrow$YOLOv8~\cite{yolov8}       & 80.1 & 79.6 & 80.4 & 78.3 \\
            FeatEnHancer~\cite{hashmi2023featenhancer}$\rightarrow$YOLOv8~\cite{yolov8} & 80.2 & 79.6 & 79.6 & 78.9 \\
            \rowcolor{myblue}
            LR-YOLOv8 (Ours)         & \textbf{82.3} & \textbf{82.5} & \textbf{83.0} & \textbf{81.6} \\
            \bottomrule
            \end{tabular}
            }
            \end{minipage}
        }{
            \captionof{table}{\textbf{Generalization on other degradation.} We assess the robustness of our method under motion blur, rain, snow, and a haze–rain mixture.}
            \label{tab:robustness_yolov8}
        }
    \end{floatrow}
\end{figure}

\vspace{-5pt}
\section{Conclusion}
\vspace{-3pt}

In this paper, we revisit the integration of image restoration and object detection under adverse conditions through the lens of Lipschitz continuity in both the input and parameter spaces. Our analysis reveals that the inherent mismatch in Lipschitz continuity between these tasks introduces instability and non-smoothness when directly cascaded. To address this, we propose a Lipschitz-regularized framework that harmonizes the two tasks by constraining the Lipschitz continuity of object detection. This is achieved through low-Lipschitz restoration learning to smooth perturbations before detection, alongside parameter-space regularization to stabilize gradient flows during training. We implement this approach as Lipschitz-Regularized YOLO (LR-YOLO), which extends to existing YOLO detectors with minimal overhead. Extensive experiments on haze and low-light benchmarks show that our method improves detection stability and optimization smoothness, contributing to more robust performance in challenging environments.

\textbf{Limitation and Future Direction.} While our method has been validated across a range of adverse conditions—including haze, rain, snow, low-light, and mixed weather—a current limitation is that each input is assumed to contain only a single type of degradation. A valuable future direction would be to extend our framework to handle inputs affected by multiple, concurrent degradations. Another promising direction is to extend our Lipschitz-continuity analysis to camouflaged object detection (COD~\cite{fan2020camouflaged,hu2023high}), as COD involves detecting objects with ambiguous, low-contrast boundaries, posing challenges similar to those in cascaded systems under adverse conditions.

\section*{Acknowledgements}

This work is supported in part by the National Natural Science Foundation of China (NSFC) under Grant Nos. 62376292, 62325605, and U21A20470; the Guangdong Provincial General Fund under Grant No. 2024A1515010208; and the Guangzhou Science and Technology Program Project under Grant Nos. 2025A04J5465 and 2024A04J6365.

\bibliographystyle{IEEEtran}
\bibliography{neurips_2025}

\clearpage
\newpage
\appendix
\section*{Appendix}

\section{Detailed Proofs}
\label{appendix:proof}

\textbf{Remark~\ref{remark:restoration}} (Lipschitz Regularization via Low-Lipschitz Restoration).~ \emph{Let:
\(
f_{\mathbf{\theta}_b, \mathbf{\theta}_d} = f_{\mathbf{\theta}_d} \circ f_{\mathbf{\theta}_b}
\)
denote the object detection model, where
\( f_{\mathbf{\theta}_b}(\cdot; \mathbf{\theta}_b) \) is the backbone network parameterized by \( \mathbf{\theta}_b \), and  
\( f_{\mathbf{\theta}_d}(\cdot; \mathbf{\theta}_d) \) is the detection head parameterized by \( \mathbf{\theta}_d \).
Similarly, let:
\(
\raisebox{0.3ex}{$g$}_{\mathbf{\theta}_b, \mathbf{\theta}_r} = f_{\mathbf{\theta}_r} \circ f_{\mathbf{\theta}_b}
\)
represent the image restoration model, where
\( f_{\mathbf{\theta}_r}(\cdot; \mathbf{\theta}_r) \) is the restoration head parameterized by \( \mathbf{\theta}_r \), sharing the same backbone \( f_{\mathbf{\theta}_b} \).
Given a weighted combination of the detection loss and the restoration loss:
\[
\mathcal{L}(\mathbf{\theta}_b, \mathbf{\theta}_d, \mathbf{\theta}_r) = \mathcal{L}_{\mathrm{det}}(f_{\mathbf{\theta}_b, \mathbf{\theta}_d}) + \lambda \cdot \mathcal{L}_{\mathrm{res}}(\raisebox{0.3ex}{$g$}_{\mathbf{\theta}_b, \mathbf{\theta}_r}), \quad \lambda > 0
\]}

\emph{
Let $\operatorname{Lip}(f_{\mathbf{\theta}_b}) := \sup_{\boldsymbol{x}} \| J_{f_{\mathbf{\theta}_b}}(\boldsymbol{x}) \|$ be the Lipschitz constant of $f_{\mathbf{\theta}_b}$ defined by jacobian norm. If: 
\begin{enumerate}[left=0pt, nosep]
    \item $\mathcal{L}_{\mathrm{res}}$ is Lipschitz continuous and $\|\nabla_{\mathbf{\theta}_b} \mathcal{L}_{\mathrm{res}}(\raisebox{0.3ex}{$g$}_{\mathbf{\theta}_b, \mathbf{\theta}_r})\| \le G$ for $G < \|\nabla_{\mathbf{\theta}_b} \mathcal{L}_{\mathrm{det}}(f_{\mathbf{\theta}_b, \mathbf{\theta}_d})\|$;
    \item There exists a training sample $\boldsymbol{x}^\star$ and $\gamma > 0$ such that: \(\left\langle \nabla_{\mathbf{\theta}_b} \| J_{f_{\mathbf{\theta}_b}}(\boldsymbol{x}^\star) \|, \nabla_{\mathbf{\theta}_b} \mathcal{L}_{\mathrm{res}}(\raisebox{0.3ex}{$g$}_{\mathbf{\theta}_b, \mathbf{\theta}_r}) \right\rangle \ge \gamma, \)
\end{enumerate}
}

\emph{
then under continuous-time gradient descent \({\mathbf{\theta}_b}^{(t+1)} \leftarrow {\mathbf{\theta}_b}^{(t)} - \mu \cdot \nabla_{{\mathbf{\theta}_b}} \mathcal{L}(\mathbf{\theta}_b, \mathbf{\theta}_d, \mathbf{\theta}_r) \) (\(\mu\) denotes the learning rate), the evolution of the Lipschitz constant satisfies:
\[
\frac{d}{dt} \left[\operatorname{Lip}(f_{\mathbf{\theta}_b})\right] \le -\lambda \cdot \gamma + \xi(t)
\]
where $\xi(t) := \left\langle \nabla_{\mathbf{\theta}_b} \| J_{f_{\mathbf{\theta}_b}}(\boldsymbol{x}^\star) \|, \nabla_{\mathbf{\theta}_b} \mathcal{L}_{\mathrm{det}}(f_{\mathbf{\theta}_b, \mathbf{\theta}_d}) \right\rangle$ is the unconstrained change induced by the detection loss and \(\gamma\) is the regularization via the restoration task. 
}

\emph{
This suggests that integrating the image restoration task directly into the detector’s feature learning by sharing the detector's backbone helps suppress the model’s sensitivity to input perturbations during training, effectively acting as a Lipschitz regularization.
}

\begin{proof}
Using Theorem 1 in~\cite{latorre2020lipschitz}, the Lipschitz constant of $f_{\theta_b}$ is:

\[
\text{Lip}(f_{\theta_b}) = \sup_{\boldsymbol{x}} \|J_{f_{\theta_b}}(\boldsymbol{x})\|
\]

Let $\boldsymbol{x}^\star$ be the input that attains or approximates this supremum. Then, during continuous-time gradient descent:

\[
\frac{d}{dt} \left[\text{Lip}(f_{\theta_b})\right] = \frac{d}{dt} \|J_{f_{\theta_b}}(\boldsymbol{x}^\star)\| = \left\langle \nabla_\theta \|J_{f_{\theta_b}}(\boldsymbol{x}^\star)\|_2, -\nabla_{\theta_b} \mathcal{L}({\theta_b},{\theta_d},{\theta_r}) \right\rangle
\]

Substituting the joint loss:

\[
\frac{d}{dt} \left[\text{Lip}(f_{\theta_b})\right] = -\left\langle \nabla_{\theta_b} \|J_{f_{\theta_b}}(\boldsymbol{x}^\star)\|, \nabla_{\theta_b} \mathcal{L}_{\mathrm{det}}(f_{\mathbf{\theta}_b, \mathbf{\theta}_d}) + \lambda \cdot \nabla_{\theta_b} \mathcal{L}_{\mathrm{res}}(\raisebox{0.3ex}{$g$}_{\mathbf{\theta}_b, \mathbf{\theta}_r}) \right\rangle
\]

Breaking into two components:

\[
\frac{d}{dt} \left[\text{Lip}(f_{\theta_b})\right] = -\left\langle \nabla_{\theta_b} \|J_{f_{\theta_b}}(\boldsymbol{x}^\star)\|, \nabla_{\theta_b} \mathcal{L}_{\mathrm{det}}(f_{\mathbf{\theta}_b, \mathbf{\theta}_d}) \right\rangle - \lambda \cdot \left\langle \nabla_{\theta_b} \|J_{f_{\theta_b}}(\boldsymbol{x}^\star)\|, \nabla_{\theta_b} \mathcal{L}_{\mathrm{res}}(\raisebox{0.3ex}{$g$}_{\mathbf{\theta}_b, \mathbf{\theta}_r}) \right\rangle
\]

By Assumption $2$, the second term is lower bounded:

\[
\left\langle \nabla_\theta \|J_{f_\theta}(\boldsymbol{x}^\star)\|_2, \nabla_\theta \mathcal{L}_{\mathrm{res}}(\raisebox{0.3ex}{$g$}_{\mathbf{\theta}_b, \mathbf{\theta}_r}) \right\rangle \ge \gamma
\]

Define the first term as $\xi(t)$, then:

\[
\frac{d}{dt} \left[\text{Lip}(f_\theta)\right] \le -\lambda \gamma + \xi(t)
\]

which completes the proof.
\end{proof}

\section{Detailed Model}
\label{appendix:model}

\begin{figure}[t!]
    \centering
    \includegraphics[width=\linewidth]{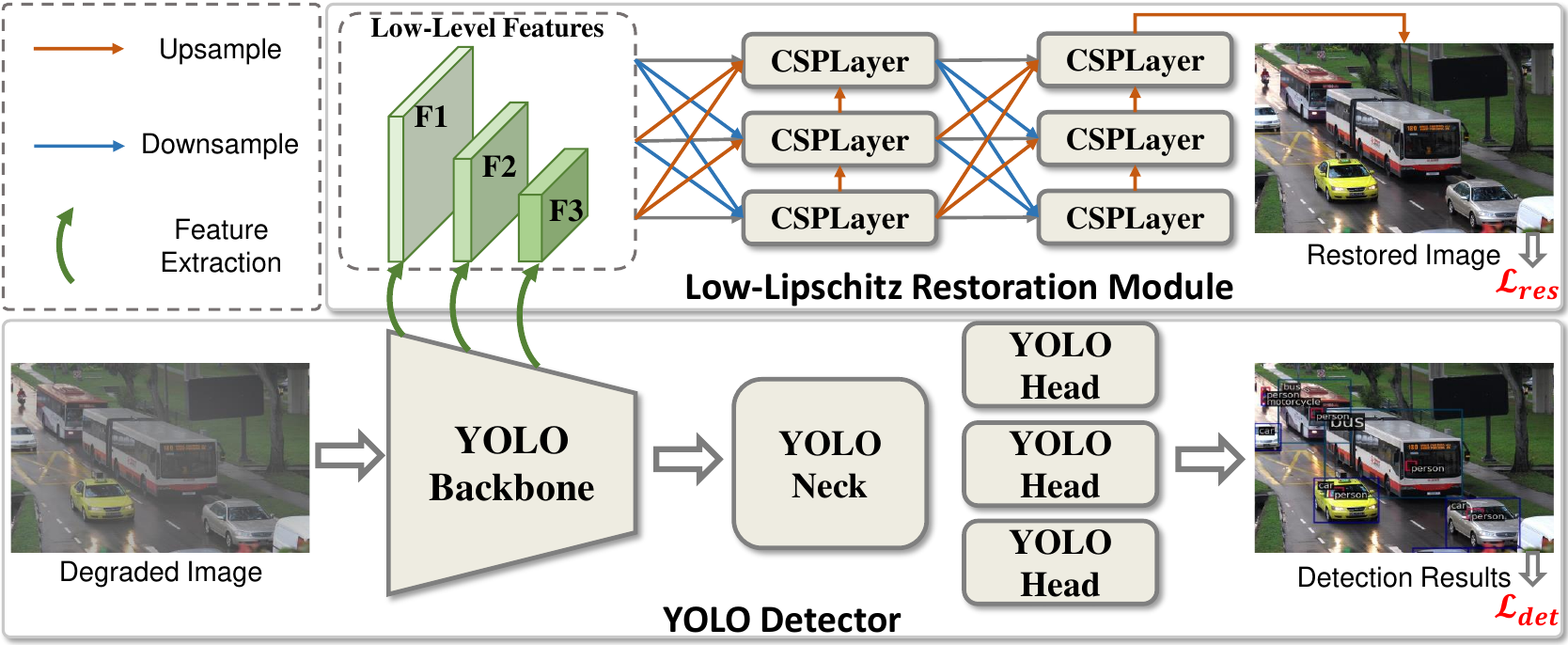}
    \caption{The overall architecture of Lipschitz-Regularized YOLO (LR-YOLO).}
    \label{fig:model}
\end{figure}

To efficiently and effectively harmonize image restoration and object detection, we integrate image restoration learning into the feature extraction process of the object detection backbone. This integration implicitly enforces Lipschitz continuity during training, thereby enhancing the stability of the detector under varying degradation intensities. As illustrated in Figure~\ref{fig:model}, we extend existing YOLO detectors by extracting low-level features from the first three stages of the backbone \emph{without modifying the original network architecture}. These features are then processed by a lightweight restoration-aware module, which reconstructs a clean version of the input image and facilitates the learning of smoother and more stable representations within the detection network.

\textbf{YOLO Detector.}
The YOLO architecture is a one-stage object detection framework that performs detection in a single forward pass, achieving high efficiency and speed. It consists of three main components, \emph{i.e.}, the Backbone, which extracts visual features from the input image; the Neck, which aggregates multi-scale features to enhance representation; and the Head, which predicts bounding boxes, class scores, and objectness. Due to its high computational efficiency and ease of deployment on edge devices, YOLO is widely used in real-time detection applications.

\textbf{Low-Lipschitz Restoration Module.} To improve the stability of object detection under adverse imaging conditions, we introduce a \emph{Restoration-Aware Module} integrated into the YOLO framework. As illustrated in Figure~\ref{fig:model}, we extract low-level features from the first three stages of the YOLO backbone (denoted as F1, F2, and F3), which preserve rich spatial and textural information essential for image restoration. Inspired by the design of the YOLO Neck and Head, these features are passed through a restoration-specific neck and decoder composed of multiple Cross Stage Partial layers (CSPLayer). The module adopts a densely connected architecture that facilitates multi-scale feature fusion, which is crucial for effective restoration learning. By progressively refining the low-level representations, the module reconstructs a restored version of the input image that is less affected by visual degradation. This restoration-aware module not only contributes to the stability of YOLO detectors during training due to the inherently smoothness of restoration, but also enhances the low-level features used by the detector for downstream detection tasks.

\section{Detailed Datasets}
\label{appendix:dataset}

\begin{table}[!t]
\tiny
\centering
\caption{Statistics of haze datasets in terms of image count and object annotations per class.}
\label{tab:dataset_stats1}
\begin{tabular}{l|ccccccc}
\toprule
Dataset & image & person & bicycle & car & bus & motorbike & Total \\
\midrule
VOC\_Haze\_Train & 8111 & 13256 & 1064 & 3267 & 822 & 1052 & 19561 \\
VOC\_Haze\_Val & 2734 & 4528 & 337  & 1201 & 213 & 325  & 6604  \\
VOC\_Haze\_Val (all objects) & 2734 & 5136 & 389 & 1528 & 254 & 367 & 7674 \\
RTTS~\cite{li2018benchmarking}     & 4322 & 7950 & 534  & 18413 & 1838 & 862 & 29577 \\
RTTS (all objects) & 4322 & 11366 & 698 & 25317 & 2590 & 1232 & 41203 \\
\bottomrule
\end{tabular}
\end{table}

\begin{table}[!t]
\tiny
\centering
\caption{Statistics of low-light datasets in terms of image count and object annotations per class.}
\label{tab:dataset_stats2}
\tiny
\begin{tabular}{l|cccccccccccc}
\toprule
\textbf{Dataset} & image & person & bicycle & car & bus & motorbike & boat & bottle & cat & chair & dog & Total \\
\midrule
VOC\_Dark\_Train & 12334 & 13256 & 1064 & 3267 & 822 & 1052 & 1140 & 1764 & 1593 & 3152 & 2025 & 29135 \\
VOC\_Dark\_Val & 3760 & 4528  & 337  & 1201 & 213 & 325  & 263  & 469  & 358  & 756  & 489  & 8939  \\
VOC\_Dark\_Val (all objects) & 3760 & 5183 & 389 & 1533 & 254 & 367 & 393 & 646 & 368 & 1268 & 529 & 10930 \\
ExDark & 2563 & 2235  & 418  & 919  & 164 & 242  & 515  & 433  & 425  & 609  & 490  & 6450  \\
\bottomrule
\end{tabular}
\end{table}

Datasets cover two challenging conditions: \emph{hazy weather} and \emph{low-light environments}. For both settings, we use real-world datasets for out-of-domain evaluation and construct synthetic training/validation sets based on PASCAL VOC~\cite{everingham2015pascal}, following IA-YOLO~\cite{liu2022image}, ReForDe~\cite{sun2022rethinking} and Vat~\cite{wu2024unsupervised}:

\emph{1) Training and Validation Data:} We construct synthetic datasets by selecting PASCAL VOC~\cite{everingham2015pascal} images containing the relevant object categories:

\begin{itemize}[leftmargin=5pt]
\item \emph{VOC\_Haze\_Train} and \emph{VOC\_Haze\_Val} consist of $8,111$ and $2,734$ images respectively. Haze is synthesized \emph{online} during training using the atmospheric scattering model with \(\beta \in [0.5, 1.5]\), while validation images are synthesized \emph{offline} once for reproducibility.
\item \emph{VOC\_Dark\_Train} and \emph{VOC\_Dark\_Val} consist of $12,334$ and $3,760$ images respectively. Low-light is simulated \emph{online} during training and \emph{offline} in validation via gamma correction with \(\gamma \in [1.5, 5]\).
\end{itemize}

\emph{2) Real-world Test Data.} We adopt \textbf{two} benchmark datasets for the out-of-domain evaluation:

\begin{itemize}[leftmargin=5pt]
\item \textbf{\emph{RTTS}}~\cite{li2018benchmarking}: contains $4,322$ real-world hazy images annotated with $5$ object categories, \emph{i.e.}, \emph{Person}, \emph{Car}, \emph{Bus}, \emph{Bicycle}, and \emph{Motorbike}.
\item \textbf{\emph{ExDark}}~\cite{loh2019getting}: contains $2,563$ real-world low-light images labeled with $10$ categories, \emph{i.e.}, \emph{People}, \emph{Car}, \emph{Bus}, \emph{Bicycle}, \emph{Motorbike}, \emph{Boat}, \emph{Bottle}, \emph{Chair}, \emph{Dog}, and \emph{Cat}.
\end{itemize}

The object annotations per class in the above datasets are presented in Table~\ref{tab:dataset_stats1} and Table~\ref{tab:dataset_stats2}.

\section{Detailed Detection Results}
\label{appendix:detection}

\begin{table}[!t]
\centering
\caption{Detailed results on \emph{VOC\_Haze\_Val} and \emph{RTTS}~\cite{li2018benchmarking}, with models trained on \emph{VOC\_Haze\_Train}.}
\label{tab:detailed_haze}
\tiny
\resizebox{\textwidth}{!}{
\begin{tabular}{l|>{\columncolor{mygray}}c>{\columncolor{white}}ccccc|>{\columncolor{mygray}}c>{\columncolor{white}}ccccc}
\hline
\multirow{2}{*}{\textbf{Methods}} 
& \multicolumn{6}{c|}{\cellcolor{myblue}\textbf{\emph{VOC\_Haze\_Val}}} 
& \multicolumn{6}{c}{\cellcolor{myblue}\textbf{\emph{RTTS}}~\cite{li2018benchmarking}} \\
& \textbf{mAP} & Person & Car & Bus & Bicycle & Motorbike 
& \textbf{mAP} & Person & Car & Bus & Bicycle & Motorbike \\
\hline
YOLOv10 & 50.5 & 55.3 & 63.5 & 46.3 & 48.1 & 39.5 & 42.6 & 70.7 & 48.9 & 20.8 & 41.0  & 31.5 \\
SFNet$\rightarrow$YOLOv10 & 77.9 & 79.2 & 82.4 & 76.5 & 76.1 & 75.3 & 45.5 & 72.2 & 53.2 & 23.9 & 42.6 & 35.4 \\
SFNet$\rightarrow$YOLOv10$^\dagger$ & 79.1 & 80.5 & 83.6 & 77.4 & 76.9 & 77.3 & 46.6 & 73.4 & 54.8 & 24.9 & 42.1  & 38.0 \\
SFNet$\rightarrow$YOLOv10$^\ddagger$ & 79.3 & 80.4 & 83.2 & 77.8 & 77.5 & 77.7 & 45.8 & 72.4 & 54.5 & 24.1 & 41.7 & 36.2 \\
ConvIR$\rightarrow$YOLOv10 & 79.9 & 80.5 & 83.7 & 79.2 & 77.8 & 78.1 & 46.1 & 72.6 & 53.9 & 24.3 & 43.7 & 36.3 \\
ConvIR$\rightarrow$YOLOv10$^\dagger$ & 80.1 & 80.9 & 84.4 & 79.2 & 77.3 & 78.6 & 46.6 & 73.2 & 55.3 & 25.0 & 42.4 & 37.0 \\
ConvIR$\rightarrow$YOLOv10$^\ddagger$ & \underline{80.5} & 81.1 & 83.9 & 80.1 & 78.3 & 79.0 & 46.5 & 72.5 & 54.7 & 24.4 & 43.7 & 37.2 \\
IA$\rightarrow$YOLOv10 & 79.9 & 79.9 & 83.8 & 76.3 & 81.4 & 78.1 & 45.4 & 72.0 & 55.4 & 22.5 & 42.3 & 34.6 \\
GDIP$\rightarrow$YOLOv10 & 79.2 & 79.3 & 83.4 & 76.8 & 79.4 & 77.4 & \underline{47.2} & 73.2 & 54.6 & 24.6 & 45.3 & 38.1 \\
FeatEnHancer$\rightarrow$YOLOv10 & 79.8 & 79.2 & 82.9 & 78.4 & 80.0 & 78.6 & 46.7 & 72.0 & 53.4 & 22.9 & 43.8 & 36.6 \\
\rowcolor{myblue}
LR-YOLOv10 (Ours) & \textbf{82.5} & 82.0 & 86.6 & 79.7 & 83.4 & 81.8 & \textbf{49.2} & 74.1 & 58.5 & 27.7 & 44.9 & 42.0 \\
\hline
YOLOv8 & 54.3 & 58.2 & 65.6 & 51.8 & 52.3 & 43.7 & 45.3 & 73.3 & 53.4 & 21.1 & 45.6 & 33.3 \\
SFNet$\rightarrow$YOLOv8 & 79.2 & 80.3 & 83.8 & 75.9 & 78.9 & 77.1 & 48.9 & 75.2 & 59.0 & 23.9 & 48.7 & 37.8 \\
SFNet$\rightarrow$YOLOv8$^\dagger$ & 80.8 & 82.0 & 85.2 & 78.2 & 79.4 & 78.9 & 49.1 & 75.6 & 59.2 & 25.2 & 47.3 & 38.0 \\
SFNet$\rightarrow$YOLOv8$^\ddagger$ & 80.3 & 81.6 & 84.8 & 77.5 & 79.4 & 78.0 & 49.3 & 75.6 & 59.8 & 24.4 & 48.3 & 38.4 \\
ConvIR$\rightarrow$YOLOv8 & 80.5 & 81.3 & 84.8 & 78.2 & 79.7 & 78.5 & 49.3 & 75.3 & 59.5 & 24.2 & 48.9 & 38.5 \\
ConvIR$\rightarrow$YOLOv8$^\dagger$ & 80.9 & 82.1 & 85.6 & 77.3 & 79.7 & 79.6 & 49.5 & 75.3 & 59.9 & 25.7 & 47.7 & 39.0 \\
ConvIR$\rightarrow$YOLOv8$^\ddagger$ & \underline{81.4} & 82.4 & 85.9 & 77.9 & 81.3 & 79.9 & 50.1 & 75.6 & 60.1 & 25.8 & 49.4 & 39.5 \\
IA$\rightarrow$YOLOv8 & 80.6 & 81.6 & 85.5 & 77.2 & 80.8 & 78.0 & 47.7 & 73.3 & 58.7 & 25.6 & 44.6 & 36.5 \\
GDIP$\rightarrow$YOLOv8 & 81.0 & 81.7 & 85.9 & 78.1 & 79.9 & 79.3 & \underline{50.3} & 75.8 & 59.7 & 27.8 & 49.9 & 38.0 \\
FeatEnHancer$\rightarrow$YOLOv8 & 81.2 & 81.4 & 85.3 & 77.0 & 81.8 & 80.4 & 48.4 & 74.8 & 57.9 & 26.1 & 47.1 & 40.7 \\
\rowcolor{myblue}
LR-YOLOv8 (Ours) & \textbf{83.3} & 83.7 & 87.6 & 79.3 & 84.4 & 82.8 & \textbf{53.2} & 76.0 & 62.9 & 30.0 & 50.2 & 46.9 \\
\hline
\end{tabular}
}

\resizebox{\textwidth}{!}{
\begin{tabular}{l|>{\columncolor{mygray}}c>{\columncolor{white}}ccccc|>{\columncolor{mygray}}c>{\columncolor{white}}ccccc}
\hline
\multirow{2}{*}{\textbf{Methods}} 
& \multicolumn{6}{c|}{\cellcolor{myblue}\textbf{\emph{VOC\_Haze\_Val (all objects)}}} 
& \multicolumn{6}{c}{\cellcolor{myblue}\textbf{\emph{RTTS (all objects)}}~\cite{li2018benchmarking}} \\
& \textbf{mAP} & Person & Car & Bus & Bicycle & Motorbike 
& \textbf{mAP} & Person & Car & Bus & Bicycle & Motorbike \\
\hline
YOLOv10 & 44.7 & 50.2 & 53.8 & 41.7 & 42.1 & 35.6 & 33.8 & 54.2 & 40.8 & 15.1 & 33.5 & 23.4 \\
SFNet$\rightarrow$YOLOv10 & 70.1 & 72.7 & 71.7 & 68.2 & 69.4 & 68.6 & 35.9 & 55.7 & 41.6 & 18.8 & 35.7 & 27.9 \\
SFNet$\rightarrow$YOLOv10$^\dagger$ & 72.1 & 75.5 & 74.4 & 69.0 & 69.5 & 71.9 & 37.1 & 58.0 & 44.6 & 19.9 & 33.9 & 28.9 \\
SFNet$\rightarrow$YOLOv10$^\ddagger$ & 71.7 & 75.1 & 73.2 & 68.9 & 69.5 & 71.4 & 36.0 & 55.9 & 42.6 & 19.1 & 34.4 & 28.2 \\
ConvIR$\rightarrow$YOLOv10 & 72.2 & 75.3 & 73.3 & 70.5 & 70.8 & 70.9 & 36.0 & 55.7 & 42.0 & 19.3 & 35.8 & 27.3 \\
ConvIR$\rightarrow$YOLOv10$^\dagger$ & \underline{72.9} & 76.2 & 75.0 & 70.6 & 70.2 & 72.5 & \underline{37.2} & 57.8 & 45.1 & 20.2 & 34.6 & 28.4 \\
ConvIR$\rightarrow$YOLOv10$^\ddagger$ & 72.6 & 76.0 & 73.7 & 70.2 & 70.8 & 72.5 & 36.5 & 55.9 & 42.8 & 19.4 & 35.3 & 29.0 \\
IA$\rightarrow$YOLOv10 & 72.0 & 74.5 & 74.0 & 68.1 & 72.8 & 70.8 & 35.8 & 55.5 & 43.4 & 18.3 & 34.3 & 27.8 \\
GDIP$\rightarrow$YOLOv10 & 70.9 & 72.7 & 72.6 & 68.1 & 70.2 & 70.7 & 37.0 & 56.1 & 42.7 & 19.7 & 37.0 & 29.2 \\
FeatEnHancer$\rightarrow$YOLOv10 & 71.6 & 72.7 & 72.7 & 69.8 & 71.6 & 71.2 & 35.8 & 55.2 & 41.7 & 18.3 & 35.6 & 28.1 \\
\rowcolor{myblue}
LR-YOLOv10 (Ours) & \textbf{74.4} & 77.5 & 76.7 & 71.3 & 75.0 & 74.6 & \textbf{38.5} & 59.1 & 45.9 & 21.3 & 37.3 & 33.0 \\
\hline
YOLOv8 & 48.3 & 52.6 & 54.3 & 45.9 & 47.9 & 40.6 & 36.2 & 56.3 & 43.3 & 17.2 & 37.5 & 26.5 \\
SFNet$\rightarrow$YOLOv8 & 71.1 & 73.9 & 73.6 & 67.5 & 70.5 & 70.1 & 38.4 & 58.1 & 46.1 & 19.5 & 39.1 & 29.1 \\
SFNet$\rightarrow$YOLOv8$^\dagger$ & 73.8 & 77.1 & 76.7 & 70.4 & 71.5 & 73.3 & 39.3 & 60.4 & 48.3 & 20.7 & 38.0 & 28.9 \\
SFNet$\rightarrow$YOLOv8$^\ddagger$ & 72.8 & 76.7 & 75.2 & 69.1 & 71.3 & 71.6 & 39.2 & 60.3 & 48.7 & 19.8 & 38.4 & 29.1 \\
ConvIR$\rightarrow$YOLOv8 & 72.8 & 76.4 & 75.0 & 69.3 & 71.5 & 71.6 & 38.7 & 58.1 & 46.4 & 19.9 & 40.0 & 29.4 \\
ConvIR$\rightarrow$YOLOv8$^\dagger$ & \underline{74.1} & 77.5 & 76.9 & 70.0 & 71.9 & 74.1 & 39.0 & 59.7 & 48.9 & 20.6 & 37.6 & 28.3 \\
ConvIR$\rightarrow$YOLOv8$^\ddagger$ & 74.0 & 77.3 & 77.0 & 69.3 & 72.3 & 74.2 & \underline{39.9} & 60.2 & 49.0 & 20.2 & 39.8 & 30.2 \\
IA$\rightarrow$YOLOv8 & 73.0 & 76.5 & 75.7 & 69.4 & 72.6 & 70.9 & 37.3 & 56.5 & 45.8 & 19.8 & 35.6 & 28.7 \\
GDIP$\rightarrow$YOLOv8 & 73.1 & 76.4 & 75.4 & 69.5 & 71.0 & 73.0 & 39.8 & 60.5 & 46.6 & 22.6 & 39.8 & 29.5 \\
FeatEnHancer$\rightarrow$YOLOv8 & 73.4 & 76.4 & 75.5 & 68.7 & 72.9 & 73.4 & 38.8 & 59.3 & 45.3 & 20.7 & 37.7 & 31.0 \\
\rowcolor{myblue}
LR-YOLOv8 (Ours) & \textbf{76.5} & 79.2 & 79.1 & 71.1 & 75.5 & 77.4 & \textbf{42.4} & 60.9 & 51.4 & 23.9 & 40.0 & 36.0 \\
\hline
\end{tabular}
}
\end{table}

We report the average precision for each category on the \emph{VOC\_Haze\_Val} and \emph{RTTS} datasets, as shown in Table~\ref{tab:detailed_haze}, and on the \emph{VOC\_Dark\_Val} and \emph{ExDark} datasets, as presented in Table~\ref{tab:detailed_low_light}.

\section{Image Restoration Evaluation}
\label{appendix:restoration}

\begin{table*}[!t]
\centering
\caption{Restoration results on \emph{VOC\_Haze\_Val} and \emph{VOC\_Dark\_Val}.}
\label{tab:restoration}
\tiny
\begin{minipage}[t]{0.49\textwidth}
\centering
\resizebox{\textwidth}{!}{
\begin{tabular}{l|cc}
\hline
\multirow{2}{*}{\textbf{Methods}} & \multicolumn{2}{c}{\cellcolor{myblue}\textbf{\emph{VOC\_Haze\_Val}}} \\ 
& LPIPS $\downarrow$ & PSNR $\uparrow$ \\
\hline
ConvIR~\cite{cui2024revitalizing}$\rightarrow$YOLOv8 & \underline{0.180} & \textbf{25.44} \\
IA~\cite{liu2022image}$\rightarrow$YOLOv8 & 0.270 & 13.12 \\
GDIP~\cite{kalwar2023gdip}$\rightarrow$YOLOv8 & 0.234 & 15.75 \\
\hline
YOLOv8 (Baseline) & 0.382 & 13.51 \\
\rowcolor{myblue}
LR-YOLOv8 (Restoration-Only) & 0.195 & \underline{23.72} \\
\rowcolor{myblue}
LR-YOLOv8 (Ours) & \textbf{0.133} & 22.72 \\
\hline
\end{tabular}
}
\end{minipage}
\hfill
\begin{minipage}[t]{0.49\textwidth}
\centering
\resizebox{\textwidth}{!}{
\begin{tabular}{l|cc}
\hline
\multirow{2}{*}{\textbf{Methods}} & \multicolumn{2}{c}{\cellcolor{myblue}\textbf{\emph{VOC\_Dark\_Val}}} \\ 
& LPIPS $\downarrow$ & PSNR $\uparrow$ \\
\hline
Retinexformer~\cite{cai2023retinexformer}$\rightarrow$YOLOv8 & 0.293 & \textbf{21.46} \\
IA~\cite{liu2022image}$\rightarrow$YOLOv8 & 0.195 & 20.73 \\
GDIP~\cite{kalwar2023gdip}$\rightarrow$YOLOv8 & \underline{0.189} & 18.82 \\
\hline
YOLOv8 (Baseline) & 0.315 & 12.00 \\
\rowcolor{myblue}
LR-YOLOv8 (Restoration-Only) & 0.245 & \underline{21.19} \\
\rowcolor{myblue}
LR-YOLOv8 (Ours) & \textbf{0.179} & 21.05 \\
\hline
\end{tabular}
}
\end{minipage}
\end{table*}

We evaluate the restoration performance of our LR-YOLO under haze and low-light conditions using peak signal-to-noise ratio (PSNR) for pixel-level fidelity and learned perceptual image patch similarity (LPIPS)~\cite{zhang2018unreasonable} for perceptual similarity. Table~\ref{tab:restoration} presents the restoration results on \emph{VOC\_Haze\_Val} and \emph{VOC\_Dark\_Val}. We compare our method with the image dehazing technique ConvIR~\cite{cui2024revitalizing}, low-light enhancement method Retinexformer~\cite{cai2023retinexformer}, and image adaptive methods IA~\cite{liu2022image} and GDIP~\cite{kalwar2023gdip}. Our full model achieves the best LPIPS score and competitive PSNR on both \emph{VOC\_Haze\_Val} and \emph{VOC\_Dark\_Val}, outperforming the baseline YOLOv8 and image adaptive pipelines. Additionally, our restoration-only variant (LR-YOLOv8 (Restoration-Only) trained only with restoration learning) achieves a balanced improvement in reconstruction fidelity.

\begin{table}[!t]
\centering
\caption{Detailed results on \emph{VOC\_Dark\_Val} and \emph{ExDark}~\cite{loh2019getting}, with models trained on \emph{VOC\_Dark\_Train}.}
\label{tab:detailed_low_light}
\tiny
\resizebox{\textwidth}{!}{
\begin{tabular}{l|>{\columncolor{mygray}}ccccccccccc}
\hline
\multirow{2}{*}{\textbf{Methods}} 
& \multicolumn{11}{c}{\cellcolor{myblue}\textbf{\emph{VOC\_Dark\_Val}}} \\
& \textbf{mAP} & Person & Car & Bus & Bicycle & Motorbike 
& Boat & Bottle & Chair & Dog & Cat \\
\hline
YOLOv10 & 62.1 & 68.7 & 72.9 & 69.0 & 70.2 & 67.1 & 56.2 & 47.0 & 43.0 & 61.7 & 65.3 \\
LLFormer$\rightarrow$YOLOv10 & 65.6 & 72.6 & 78.0 & 71.8 & 74.0 & 70.7 & 59.0 & 46.9 & 46.2 & 66.1 & 70.3 \\
LLFormer$\rightarrow$YOLOv10$^\dagger$ & 64.7 & 73.9 & 77.4 & 69.7 & 73.5 & 71.7 & 56.4 & 47.2 & 43.1 & 65.0 & 68.6 \\
LLFormer$\rightarrow$YOLOv10$^\ddagger$ & 66.3 & 73.2 & 78.2 & 71.8 & 73.7 & 71.7 & 60.1 & 50.0 & 46.0 & 67.5 & 70.6 \\
RetinexFormer$\rightarrow$YOLOv10 & 66.3 & 73.3 & 78.8 & 73.6 & 72.9 & 70.7 & 59.8 & 49.7 & 46.0 & 66.4 & 71.7 \\
RetinexFormer$\rightarrow$YOLOv10$^\dagger$ & 66.0 & 74.8 & 78.5 & 72.7 & 75.4 & 72.7 & 57.2 & 49.5 & 45.2 & 64.9 & 69.2 \\
RetinexFormer$\rightarrow$YOLOv10$^\ddagger$ & 66.9 & 74.9 & 78.6 & 72.4 & 75.2 & 72.6 & 60.3 & 49.6 & 46.5 & 67.4 & 71.8 \\
IA$\rightarrow$YOLOv10 & 66.0 & 73.1 & 78.5 & 68.2 & 74.5 & 69.4 & 60.6 & 52.0 & 48.2 & 67.1 & 68.0 \\
GDIP$\rightarrow$YOLOv10 & 65.8 & 74.3 & 77.9 & 71.2 & 77.4 & 72.5 & 56.5 & 49.4 & 45.4 & 67.0 & 66.8 \\
FeatEnHancer$\rightarrow$YOLOv10 & \underline{67.6} & 75.3 & 79.1 & 73.6 & 76.7 & 73.8 & 59.6 & 52.3 & 49.2 & 68.3 & 68.6 \\
\rowcolor{myblue}
LR-YOLOv10 (Ours) & \textbf{70.6} & 77.6 & 82.5 & 76.3 & 78.3 & 75.9 & 65.6 & 56.3 & 52.1 & 69.3 & 72.5 \\
\hline
YOLOv8 & 63.4 & 70.2 & 76.1 & 69.3 & 72.1 & 67.4 & 59.9 & 45.9 & 44.2 & 63.4 & 65.9 \\
LLFormer$\rightarrow$YOLOv8 & 66.2 & 74.5 & 80.0 & 71.5 & 74.3 & 71.7 & 59.0 & 45.4 & 47.2 & 67.9 & 70.3 \\
LLFormer$\rightarrow$YOLOv8$^\dagger$ & 66.2 & 76.3 & 80.6 & 71.9 & 75.1 & 71.7 & 59.2 & 45.8 & 46.2 & 65.3 & 70.0 \\
LLFormer$\rightarrow$YOLOv8$^\ddagger$ & 66.2 & 75.0 & 80.7 & 71.6 & 75.7 & 72.7 & 59.5 & 47.4 & 45.3 & 65.5 & 68.4 \\
RetinexFormer$\rightarrow$YOLOv8 & 67.8 & 76.5 & 81.1 & 73.5 & 73.8 & 71.0 & 62.9 & 49.8 & 47.7 & 69.4 & 72.2 \\
RetinexFormer$\rightarrow$YOLOv8$^\dagger$ & 67.7 & 77.5 & 80.9 & 72.8 & 74.8 & 73.6 & 61.6 & 48.6 & 49.6 & 67.3 & 70.2 \\
RetinexFormer$\rightarrow$YOLOv8$^\ddagger$ & 68.6 & 77.4 & 81.6 & 75.2 & 75.2 & 74.3 & 61.3 & 51.0 & 49.4 & 68.5 & 71.8 \\
IA$\rightarrow$YOLOv8 & 66.5 & 73.2 & 79.8 & 70.7 & 73.1 & 73.0 & 61.6 & 50.1 & 48.6 & 66.5 & 68.4 \\
GDIP$\rightarrow$YOLOv8 & \underline{68.9} & 77.1 & 81.6 & 73.1 & 76.2 & 75.0 & 61.9 & 51.4 & 49.9 & 69.2 & 73.3 \\
FeatEnHancer$\rightarrow$YOLOv8 & 68.7 & 75.8 & 79.7 & 73.2 & 76.9 & 76.7 & 60.6 & 52.5 & 51.0 & 69.6 & 71.3 \\
\rowcolor{myblue}
LR-YOLOv8 (Ours) & \textbf{71.7} & 78.5 & 82.9 & 77.2 & 79.7 & 78.3 & 62.9 & 60.1 & 52.6 & 70.3 & 74.3 \\
\hline
\end{tabular}
}

\resizebox{\textwidth}{!}{
\begin{tabular}{l|>{\columncolor{mygray}}ccccccccccc}
\hline
\multirow{2}{*}{\textbf{Methods}} 
& \multicolumn{11}{c}{\cellcolor{myblue}\textbf{\emph{VOC\_Dark\_Val (all objects)}}} \\
& \textbf{mAP} & Person & Car & Bus & Bicycle & Motorbike 
& Boat & Bottle & Chair & Dog & Cat \\
\hline
YOLOv10 & 55.0 & 62.3 & 64.2 & 61.7 & 62.9 & 60.9 & 42.9 & 37.5 & 34.4 & 58.7 & 64.2 \\
LLFormer$\rightarrow$YOLOv10 & 58.0 & 67.3 & 68.1 & 64.7 & 66.3 & 64.1 & 45.3 & 37.5 & 36.9 & 62.1 & 67.6 \\
LLFormer$\rightarrow$YOLOv10$^\dagger$ & 57.5 & 68.0 & 67.9 & 63.8 & 65.3 & 64.9 & 43.4 & 37.9 & 34.7 & 61.5 & 67.4 \\
LLFormer$\rightarrow$YOLOv10$^\ddagger$ & 59.2 & 70.3 & 71.1 & 64.4 & 67.7 & 67.2 & 46.9 & 37.8 & 63.5 & 63.0 & 67.4 \\
RetinexFormer$\rightarrow$YOLOv10 & 58.6 & 68.3 & 68.5 & 65.7 & 64.9 & 63.3 & 45.9 & 40.0 & 36.8 & 62.9 & 70.1 \\
RetinexFormer$\rightarrow$YOLOv10$^\dagger$ & 58.4 & 68.9 & 68.3 & 65.0 & 66.8 & 66.5 & 43.7 & 40.1 & 36.9 & 61.0 & 67.1 \\
RetinexFormer$\rightarrow$YOLOv10$^\ddagger$ & 59.2 & 68.8 & 68.7 & 65.5 & 67.8 & 65.8 & 46.6 & 39.5 & 37.3 & 63.4 & 69.1 \\
IA$\rightarrow$YOLOv10 & 58.7 & 68.5 & 68.8 & 62.0 & 67.8 & 64.1 & 47.7 & 41.9 & 67.9 & 63.2 & 64.9 \\
GDIP$\rightarrow$YOLOv10 & 58.5 & 68.5 & 68.0 & 64.1 & 69.1 & 67.1 & 44.4 & 39.3 & 36.5 & 62.9 & 65.2 \\
FeatEnHancer$\rightarrow$YOLOv10 & \underline{59.9} & 69.1 & 69.0 & 65.5 & 68.5 & 68.2 & 45.7 & 41.1 & 40.0 & 64.2 & 67.3 \\
\rowcolor{myblue}
LR-YOLOv10 (Ours) & \textbf{62.7} & 71.5 & 72.4 & 68.1 & 71.1 & 70.1 & 50.1 & 45.6 & 41.3 & 66.2 & 71.1 \\
\hline
YOLOv8 & 55.8 & 63.8 & 66.2 & 61.5 & 64.0 & 61.5 & 46.1 & 36.7 & 33.9 & 60.6 & 64.3 \\
LLFormer$\rightarrow$YOLOv8 & 58.7 & 69.6 & 70.1 & 64.1 & 67.3 & 65.5 & 46.3 & 36.1 & 36.6 & 63.9 & 67.8 \\
LLFormer$\rightarrow$YOLOv8$^\dagger$ & 58.8 & 70.3 & 71.0 & 64.6 & 67.2 & 66.1 & 45.7 & 36.9 & 36.6 & 61.9 & 67.6 \\
LLFormer$\rightarrow$YOLOv8$^\ddagger$ & 59.2 & 70.3 & 71.1 & 64.4 & 67.7 & 67.2 & 46.9 & 37.8 & 36.5 & 63.0 & 67.4 \\
RetinexFormer$\rightarrow$YOLOv8 & 59.5 & 70.2 & 70.9 & 66.0 & 65.3 & 64.5 & 48.4 & 38.6 & 37.2 & 64.5 & 69.4 \\
RetinexFormer$\rightarrow$YOLOv8$^\dagger$ & 60.0 & 71.5 & 71.5 & 65.1 & 67.4 & 67.8 & 47.5 & 38.7 & 38.5 & 62.8 & 69.0 \\
RetinexFormer$\rightarrow$YOLOv8$^\ddagger$ & 61.0 & 71.4 & 72.2 & 67.1 & 69.1 & 68.8 & 48.3 & 40.9 & 38.6 & 64.6 & 69.4 \\
IA$\rightarrow$YOLOv8 & 59.2 & 68.5 & 70.5 & 64.7 & 65.4 & 66.9 & 48.0 & 40.1 & 38.4 & 62.9 & 66.9 \\
GDIP$\rightarrow$YOLOv8 & \underline{61.2} & 71.3 & 71.9 & 65.1 & 68.6 & 68.4 & 49.2 & 41.9 & 39.5 & 65.0 & 71.5 \\
FeatEnHancer$\rightarrow$YOLOv8 & 60.8 & 69.7 & 69.9 & 65.6 & 69.3 & 70.0 & 47.0 & 41.9 & 40.7 & 65.4 & 68.9 \\
\rowcolor{myblue}
LR-YOLOv8 (Ours) & \textbf{63.9} & 72.3 & 73.3 & 70.1 & 72.3 & 71.6 & 49.0 & 48.2 & 42.3 & 67.1 & 72.7 \\
\hline
\end{tabular}
}

\resizebox{\textwidth}{!}{
\begin{tabular}{l|>{\columncolor{mygray}}ccccccccccc}
\hline
\multirow{2}{*}{\textbf{Methods}} 
& \multicolumn{11}{c}{\cellcolor{myblue}\textbf{\emph{ExDark}}~\cite{loh2019getting}} \\
& \textbf{mAP} & Person & Car & Bus & Bicycle & Motorbike 
& Boat & Bottle & Chair & Dog & Cat \\
\hline 
YOLOv10 & 49.2 & 53.1 & 50.0 & 63.8 & 58.9 & 34.0 & 43.8 & 46.3 & 43.6 & 53.8 & 44.8 \\
LLFormer$\rightarrow$YOLOv10 & 46.3 & 49.3 & 45.6 & 56.2 & 60.5 & 34.2 & 41.9 & 42.3 & 40.3 & 50.9 & 42.0 \\
LLFormer$\rightarrow$YOLOv10$^\dagger$ & 47.0 & 51.7 & 46.8 & 58.1 & 62.6 & 37.4 & 40.5 & 43.2 & 38.9 & 50.2 & 41.0 \\
LLFormer$\rightarrow$YOLOv10$^\ddagger$ & 49.5 & 54.3 & 47.9 & 60.0 & 63.9 & 37.0 & 44.3 & 46.0 & 42.0 & 53.9 & 45.6 \\
RetinexFormer$\rightarrow$YOLOv10 & 47.6 & 50.4 & 46.8 & 59.3 & 58.5 & 34.4 & 43.6 & 44.2 & 42.0 & 52.7 & 44.4 \\
RetinexFormer$\rightarrow$YOLOv10$^\dagger$ & 45.8 & 49.8 & 46.9 & 55.6 & 60.9 & 35.3 & 41.6 & 42.9 & 39.0 & 47.7 & 38.2 \\
RetinexFormer$\rightarrow$YOLOv10$^\ddagger$ & 47.5 & 52.2 & 46.7 & 57.0 & 61.4 & 35.1 & 43.4 & 44.0 & 40.8 & 51.6 & 42.9 \\
IA$\rightarrow$YOLOv10 & 50.4 & 55.9 & 50.1 & 64.2 & 61.9 & 34.5 & 45.5 & 48.5 & 43.1 & 54.1 & 46.2 \\
GDIP$\rightarrow$YOLOv10 & 48.9 & 51.7 & 47.8 & 63.4 & 61.1 & 36.7 & 42.3 & 43.7 & 39.1 & 55.5 & 47.9 \\
FeatEnHancer$\rightarrow$YOLOv10 & \underline{50.9} & 54.0 & 48.9 & 65.9 & 61.8 & 37.1 & 45.4 & 45.4 & 43.0 & 58.3 & 49.4 \\
\rowcolor{myblue} LR-YOLOv10 (Ours) & \textbf{53.8} & 57.2 & 53.7 & 69.0 & 66.3 & 40.3 & 44.3 & 53.3 & 44.7 & 59.6 & 49.4 \\
\hline 
YOLOv8 & 50.0 & 55.5 & 52.4 & 63.3 & 58.9 & 31.8 & 42.0 & 48.4 & 42.2 & 58.0 & 47.5 \\
LLFormer$\rightarrow$YOLOv8 & 46.6 & 50.7 & 47.6 & 55.2 & 58.0 & 30.5 & 41.0 & 48.2 & 38.8 & 52.5 & 43.7 \\
LLFormer$\rightarrow$YOLOv8$^\dagger$ & 47.9 & 53.5 & 50.6 & 57.9 & 60.7 & 32.7 & 39.6 & 49.3 & 38.4 & 52.2 & 43.8 \\
LLFormer$\rightarrow$YOLOv8$^\ddagger$ & 48.6 & 53.9 & 49.6 & 57.9 & 60.9 & 33.3 & 41.7 & 50.0 & 40.0 & 52.6 & 45.9 \\
RetinexFormer$\rightarrow$YOLOv8 & 47.6 & 51.3 & 49.4 & 58.1 & 57.3 & 31.4 & 41.6 & 50.6 & 39.2 & 54.1 & 43.2 \\
RetinexFormer$\rightarrow$YOLOv8$^\dagger$ & 49.5 & 54.5 & 51.6 & 60.6 & 60.8 & 34.6 & 42.5 & 50.6 & 40.0 & 53.3 & 46.5 \\
RetinexFormer$\rightarrow$YOLOv8$^\ddagger$ & 49.5 & 55.1 & 50.2 & 58.5 & 60.7 & 34.4 & 41.9 & 51.0 & 42.6 & 55.0 & 45.8 \\
IA$\rightarrow$YOLOv8 & 49.6 & 56.5 & 52.0 & 60.0 & 60.8 & 32.1 & 43.0 & 48.9 & 41.5 & 55.8 & 45.1 \\
GDIP$\rightarrow$YOLOv8 & 51.2 & 57.1 & 51.0 & 61.7 & 65.1 & 36.0 & 44.8 & 47.7 & 43.6 & 57.1 & 48.4 \\
FeatEnHancer$\rightarrow$YOLOv8 & \underline{51.8} & 55.5 & 50.0 & 63.6 & 61.7 & 36.5 & 43.5 & 51.1 & 45.7 & 59.5 & 50.9 \\
\rowcolor{myblue} LR-YOLOv8 (Ours) & \textbf{54.5} & 60.2 & 56.3 & 66.9 & 66.6 & 37.9 & 43.7 & 54.9 & 44.8 & 61.6 & 52.1 \\
\hline 
\end{tabular}
}
\end{table}

\section{Qualitative Comparison}
\label{appendix:qualitative}

\begin{figure}[t!]
    \centering
    \includegraphics[width=0.85\linewidth]{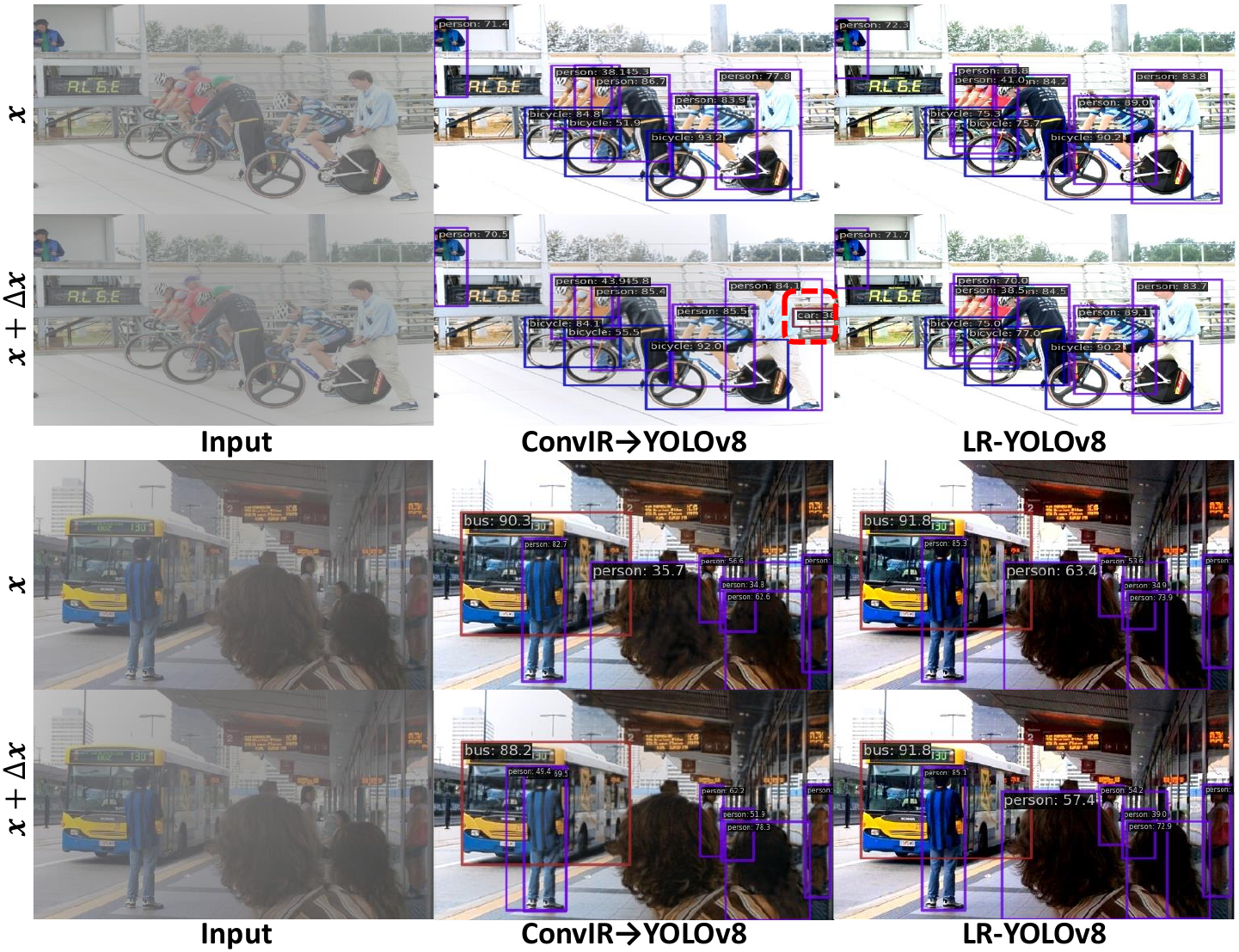}
    \caption{Qualitative comparison of detection result stability between the cascade method (ConvIR$\rightarrow$YOLOv8) and our LR-YOLOv8.}
    \label{fig:stability_example}
\end{figure}

\begin{figure}[t!]
    \centering
    \includegraphics[width=\linewidth]{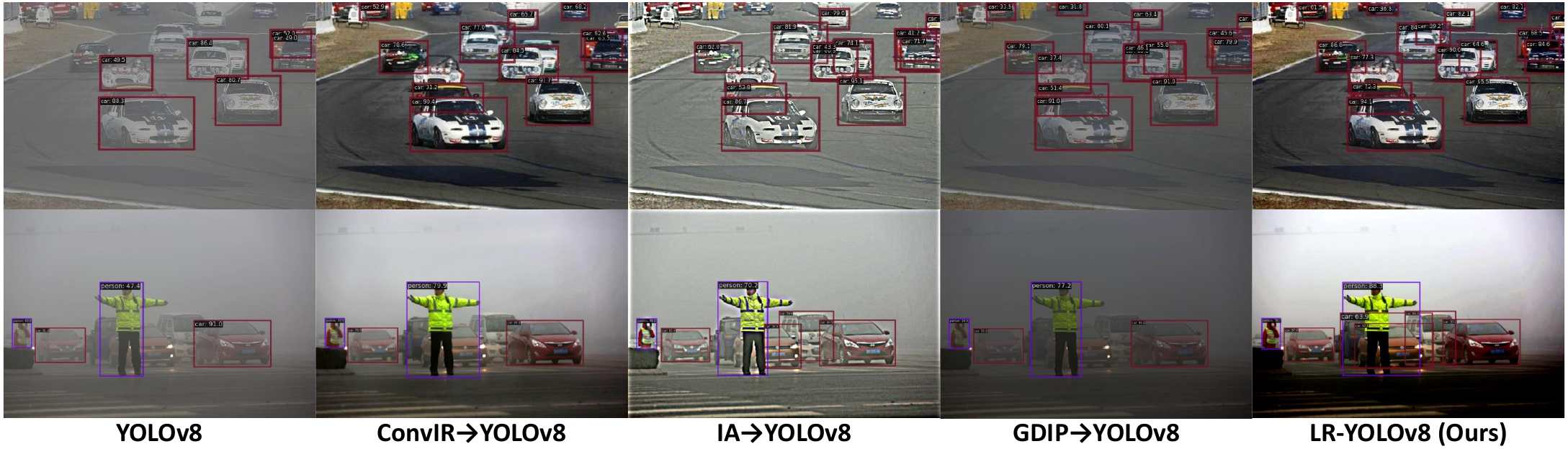}
    \caption{Qualitative comparisons on both \emph{VOC\_Haze\_Val} and \emph{RTTS} between our LR-YOLOv8 and other methods.}
    \label{fig:qualitative_comparison1}
\end{figure}

\begin{figure}[t!]
    \centering
    \includegraphics[width=\linewidth]{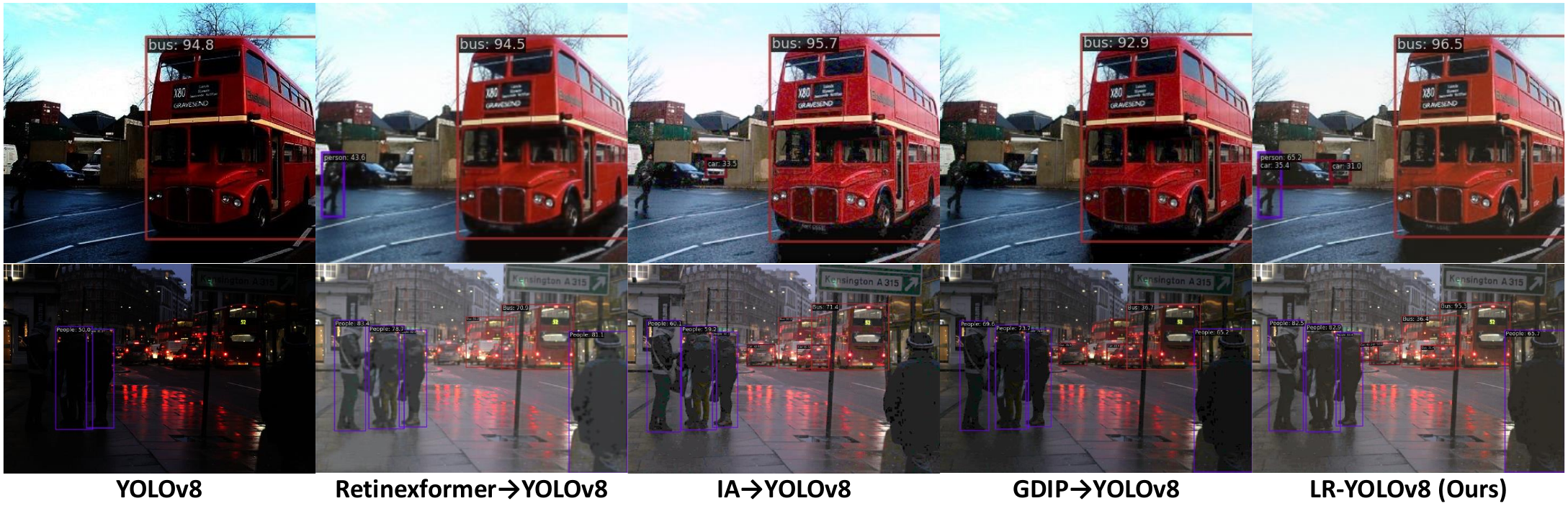}
    \caption{Qualitative comparisons on both \emph{VOC\_Dark\_Val} and \emph{ExDark} between our LR-YOLOv8 and other methods.}
    \label{fig:qualitative_comparison2}
\end{figure}

Figure~\ref{fig:lipschitz} (a) illustrates that the detector features of the cascade method are highly sensitive to minor haze density variations \(\Delta \boldsymbol{x}\), while our Lipschitz-regularized framework maintains stability. A qualitative comparison of detection result stability is presented in Figure~\ref{fig:stability_example}. When two haze inputs \(\boldsymbol{x}\) and \(\boldsymbol{x} + \Delta \boldsymbol{x}\) with slight haze density variations \(\Delta \boldsymbol{x}\) are fed into the model, the detection results of the cascade method (ConvIR$\rightarrow$YOLOv8) exhibits significant instability even though those haze can be mitigated by the image dehazing method. For example, a car is detected in one case but not in another, and a person is suddenly undetected. This highlights the instability inherent in the cascade framework. Visual examples in Figure~\ref{fig:qualitative_comparison1} (haze condition) and Figure~\ref{fig:qualitative_comparison2} (low-light condition) qualitatively illustrate the effectiveness of our method in improving detection accuracy and perceptual quality, thereby enhancing human trust in detection results.

\clearpage

\section{Broader Impacts}
\label{appendix:broader}

Improving object detection in adverse weather and low-light environments has significant implications for safety-critical applications, such as autonomous driving, traffic surveillance, and search-and-rescue missions. In particular, autonomous vehicles often operate under unpredictable environmental conditions. Failure to accurately detect pedestrians, vehicles, or obstacles in foggy or nighttime scenarios can lead to life-threatening consequences. Our method aims to fill this gap by jointly enhancing visual clarity and detection accuracy, offering a potential safety upgrade to existing perception pipelines.

Nevertheless, this line of research also entails broader considerations. First, the deployment of advanced visual detection systems could increase surveillance capabilities in urban and rural areas. While this may improve public security, it also raises concerns about privacy and the potential for misuse by authoritarian entities.

Second, performance across different demographic and geographic contexts should be evaluated. Adverse weather conditions may vary significantly between regions (e.g., smog vs. marine fog), and ensuring that models generalize fairly across different environments and communities is crucial to avoid biased deployment outcomes.

Lastly, we acknowledge that improved detection in low-visibility environments might be repurposed for military or security applications. While the proposed method is designed for civilian safety and transportation enhancement, dual-use risks exist.

\clearpage
\newpage
\section*{NeurIPS Paper Checklist}

The checklist is designed to encourage best practices for responsible machine learning research, addressing issues of reproducibility, transparency, research ethics, and societal impact. Do not remove the checklist: {\bf The papers not including the checklist will be desk rejected.} The checklist should follow the references and follow the (optional) supplemental material.  The checklist does NOT count towards the page
limit. 

Please read the checklist guidelines carefully for information on how to answer these questions. For each question in the checklist:
\begin{itemize}
    \item You should answer \answerYes{}, \answerNo{}, or \answerNA{}.
    \item \answerNA{} means either that the question is Not Applicable for that particular paper or the relevant information is Not Available.
    \item Please provide a short (1–2 sentence) justification right after your answer (even for NA). 
\end{itemize}

{\bf The checklist answers are an integral part of your paper submission.} They are visible to the reviewers, area chairs, senior area chairs, and ethics reviewers. You will be asked to also include it (after eventual revisions) with the final version of your paper, and its final version will be published with the paper.

The reviewers of your paper will be asked to use the checklist as one of the factors in their evaluation. While "\answerYes{}" is generally preferable to "\answerNo{}", it is perfectly acceptable to answer "\answerNo{}" provided a proper justification is given (e.g., "error bars are not reported because it would be too computationally expensive" or "we were unable to find the license for the dataset we used"). In general, answering "\answerNo{}" or "\answerNA{}" is not grounds for rejection. While the questions are phrased in a binary way, we acknowledge that the true answer is often more nuanced, so please just use your best judgment and write a justification to elaborate. All supporting evidence can appear either in the main paper or the supplemental material, provided in appendix. If you answer \answerYes{} to a question, in the justification please point to the section(s) where related material for the question can be found.

IMPORTANT, please:
\begin{itemize}
    \item {\bf Delete this instruction block, but keep the section heading ``NeurIPS Paper Checklist"},
    \item  {\bf Keep the checklist subsection headings, questions/answers and guidelines below.}
    \item {\bf Do not modify the questions and only use the provided macros for your answers}.
\end{itemize}

\begin{enumerate}

\item {\bf Claims}
    \item[] Question: Do the main claims made in the abstract and introduction accurately reflect the paper's contributions and scope?
    \item[] Answer: \answerYes{} %
    \item[] Justification: The abstract and introduction clearly state the claims made.

\item {\bf Limitations}
    \item[] Question: Does the paper discuss the limitations of the work performed by the authors?
    \item[] Answer: \answerYes{} %
    \item[] Justification: The paper includes a section discussing limitations.

\item {\bf Theory assumptions and proofs}
    \item[] Question: For each theoretical result, does the paper provide the full set of assumptions and a complete (and correct) proof?
    \item[] Answer: \answerYes{} %
    \item[] Justification: This paper provides the proof in the appendix.

    \item {\bf Experimental result reproducibility}
    \item[] Question: Does the paper fully disclose all the information needed to reproduce the main experimental results of the paper to the extent that it affects the main claims and/or conclusions of the paper (regardless of whether the code and data are provided or not)?
    \item[] Answer: \answerYes{} %
    \item[] Justification: This paper provides PyTorch-like codes for reproducibility in the appendix.

\item {\bf Open access to data and code}
    \item[] Question: Does the paper provide open access to the data and code, with sufficient instructions to faithfully reproduce the main experimental results, as described in supplemental material?
    \item[] Answer: \answerNo{} %
    \item[] Justification: This paper does not provide open access to data and code.

\item {\bf Experimental setting/details}
    \item[] Question: Does the paper specify all the training and test details (e.g., data splits, hyperparameters, how they were chosen, type of optimizer, etc.) necessary to understand the results?
    \item[] Answer: \answerYes{} %
    \item[] Justification: This paper specifies all relevant training and test details.

\item {\bf Experiment statistical significance}
    \item[] Question: Does the paper report error bars suitably and correctly defined or other appropriate information about the statistical significance of the experiments?
    \item[] Answer: \answerNo{} %
    \item[] Justification: This paper does not report error bars.

\item {\bf Experiments compute resources}
    \item[] Question: For each experiment, does the paper provide sufficient information on the computer resources (type of compute workers, memory, time of execution) needed to reproduce the experiments?
    \item[] Answer: \answerYes{} %
    \item[] Justification: This paper provides details on the type of compute workers (GPUs), memory requirements, and time of execution for each experiment, ensuring reproducibility.
    
\item {\bf Code of ethics}
    \item[] Question: Does the research conducted in the paper conform, in every respect, with the NeurIPS Code of Ethics \url{https://neurips.cc/public/EthicsGuidelines}?
    \item[] Answer: \answerYes{} %
    \item[] Justification: The research adheres to the NeurIPS Code of Ethics.

\item {\bf Broader impacts}
    \item[] Question: Does the paper discuss both potential positive societal impacts and negative societal impacts of the work performed?
    \item[] Answer: \answerYes{} %
    \item[] Justification: The paper includes a section discussing potential positive impacts.
    
\item {\bf Safeguards}
    \item[] Question: Does the paper describe safeguards that have been put in place for responsible release of data or models that have a high risk for misuse (e.g., pretrained language models, image generators, or scraped datasets)?
    \item[] Answer: \answerNA{} %
    \item[] Justification: The paper poses no such risks.

\item {\bf Licenses for existing assets}
    \item[] Question: Are the creators or original owners of assets (e.g., code, data, models), used in the paper, properly credited and are the license and terms of use explicitly mentioned and properly respected?
    \item[] Answer: \answerYes{} %
    \item[] Justification: The paper properly credits the creators and original owners of the assets used, and the licenses and terms of use are explicitly mentioned and respected.

\item {\bf New assets}
    \item[] Question: Are new assets introduced in the paper well documented and is the documentation provided alongside the assets?
    \item[] Answer: \answerNA{} %
    \item[] Justification: This paper does not release new assets

\item {\bf Crowdsourcing and research with human subjects}
    \item[] Question: For crowdsourcing experiments and research with human subjects, does the paper include the full text of instructions given to participants and screenshots, if applicable, as well as details about compensation (if any)? 
    \item[] Answer: \answerNA{} %
    \item[] Justification: The paper does not include experiments and research with human subjects.

\item {\bf Institutional review board (IRB) approvals or equivalent for research with human subjects}
    \item[] Question: Does the paper describe potential risks incurred by study participants, whether such risks were disclosed to the subjects, and whether Institutional Review Board (IRB) approvals (or an equivalent approval/review based on the requirements of your country or institution) were obtained?
    \item[] Answer: \answerYes{} %
    \item[] Justification: The paper describes the potential risks to study participants, confirms that these risks were disclosed, and states that IRB approval was obtained for the user study.

\item {\bf Declaration of LLM usage}
    \item[] Question: Does the paper describe the usage of LLMs if it is an important, original, or non-standard component of the core methods in this research? Note that if the LLM is used only for writing, editing, or formatting purposes and does not impact the core methodology, scientific rigorousness, or originality of the research, declaration is not required.
    \item[] Answer: \answerNA{} %
    \item[] Justification: The core method development in this research does not involve LLMs as any important, original, or non-standard components.

\end{enumerate}

\end{document}